%% file: _main.tex
\input{sections/_constants}
\arxiv

\pdfoutput=1
\documentclass[10pt,twocolumn,letterpaper]{article}
\input{meta/cvpr_header}

\newcommand{\algname}{{AlignPrune}}

\begin{document}
\title{\paperTitle}
\author{\authorBlock}
\maketitle

\freefootnote{
\hspace{-12pt}
\textsuperscript{\Letter}Corresponding author.
}
\makeatother

\input{sections/00_abstract}
\input{sections/01_intro}
\input{sections/02_related}

\input{sections/03_method}

\input{sections/04_experiment}

\input{sections/10_conclusion}

\pagebreak

\nbf{Acknowledgments}
This research is supported by 
the EDB Space Technology Development Programme under Project S22-19016-STDP.
The authors sincerely thank the anonymous Reviewers and Area Chairs for their valuable feedback and constructive comments.

{\small
\bibliographystyle{meta/ieeenat_fullname}
\bibliography{sections/11_references}
}

\appendix \input{sections/12_appendix}

\end{document}

%% file: sections/_constants.tex
\def\paperID{44661}

\def\confName{CVPR}
\def\confYear{2026}

\def\paperTitle{Beyond Loss Values: Robust Dynamic Pruning via Loss Trajectory Alignment}

\def\authorBlock{
{\fontsize{12}{16}\selectfont
    Huaiyuan Qin\textsuperscript{1}%
\qquad
    Muli Yang\textsuperscript{1}%
\qquad
    Gabriel James Goenawan\textsuperscript{1}%
\qquad 
    Kai Wang\textsuperscript{2}%
}\\
{\fontsize{12}{16}\selectfont
    Zheng Wang\textsuperscript{3}%
\qquad
    Peng Hu\textsuperscript{4}%
\qquad 
    Xi Peng\textsuperscript{4}%
\qquad
    Hongyuan Zhu\textsuperscript{1\Letter}%
}\vspace{2.5pt}\\

{\fontsize{10.8}{16}\selectfont
\textsuperscript{1}Institute for Infocomm Research (I\textsuperscript{2}R), A*STAR, Singapore
}\\

{\fontsize{10.8}{16}\selectfont\textsuperscript{2}National University of Singapore
}\hspace{9pt}
{\fontsize{10.9}{16}\selectfont\textsuperscript{3}Wuhan University}
\hspace{9pt}
{\fontsize{10.9}{16}\selectfont\textsuperscript{4}Sichuan University
}\\

{\tt\fontsize{9.6}{16}\selectfont
\{qinhy, yangml, goenawan, zhuh\}@a-star.edu.sg,
kai.wang@comp.nus.edu.sg,\hspace{-6.28pt}}\\

{\tt\fontsize{9.6}{16}\selectfont
wangzwhu@whu.edu.cn,
\{penghu.ml, pengx.gm\}@gmail.com\hspace{-6.28pt}
}
}

\newif\ifreview 
\newif\ifarxiv \newcommand{\arxiv}{\arxivtrue}
\newif\ifcamera 
\newif\ifrebuttal 

%% file: meta/cvpr_header.tex
\ifreview \usepackage[review]{meta/cvpr} \fi
\ifarxiv \usepackage[pagenumbers]{meta/cvpr} \fi
\ifrebuttal \usepackage[rebuttal]{meta/cvpr} \fi
\ifcamera \usepackage{meta/cvpr} \fi

\input{sections/_macros}  

\usepackage{xr-hyper}

\makeatletter
\newcommand*{\addFileDependency}[1]{
  \typeout{(#1)}
  \@addtofilelist{#1}
  \IfFileExists{#1}{}{\typeout{No file #1.}}
}

\makeatother

\definecolor{cvprblue}{rgb}{0.21,0.49,0.74}
\usepackage[pagebackref,breaklinks,colorlinks,allcolors=cvprblue]{hyperref}
\usepackage[capitalize]{cleveref}
\crefname{section}{Sec.}{Secs.}
\crefname{table}{Table}{Tables}
\crefname{figure}{Fig.}{Figs.}

%% file: sections/_macros.tex
\usepackage{fix-cm}
\usepackage{array}
\usepackage{bbm}
\usepackage{nicematrix}

\usepackage{tipa}
\usepackage{dsfont}
\usepackage{etoolbox}  

\usepackage[noend]{algorithmic}
\usepackage{algorithm}

\usepackage{float}
\usepackage{newfloat}
\usepackage{listings}
\floatstyle{ruled}
\newfloat{listing}{tb}{lst}{}
\floatname{listing}{\small Algorithm}

\definecolor{mygray}{RGB}{234,234,234}

\usepackage{adjustbox}
\usepackage[table,dvipsnames]{xcolor}

\definecolor{darkgreen}{rgb}{0.13, 0.55, 0.13}

\usepackage{graphicx}	
\usepackage{amsmath}	
\usepackage{amssymb}	
\usepackage{booktabs}
\usepackage{times}
\usepackage{epsfig}
\usepackage{caption}
\usepackage{float}
\usepackage{placeins}
\usepackage{color, colortbl}
\usepackage{stfloats}
\usepackage{enumitem}
\usepackage{tabularx}
\usepackage{xstring}
\usepackage{multirow}
\usepackage{xspace}
\usepackage{url}
\usepackage{subcaption}
\usepackage{xcolor}

\ifcamera \usepackage[accsupp]{axessibility} \fi

% ...

\newcommand{\nbf}[1]{\paragraph{#1.}}

\ifarxiv  \fi

\newcommand{\R}[1]{{%
    \textbf{%
        \ifstrequal{#1}{1}{\textcolor{red}{R#1}}{%
        \ifstrequal{#1}{2}{\textcolor{blue}{R#1}}{%
        \ifstrequal{#1}{3}{\textcolor{magenta}{R#1}}{%
        \ifstrequal{#1}{4}{\textcolor{teal}{R#1}}{%
                           \textcolor{cyan}{R#1}%
        }}}}%
    }%
}}

\DeclareMathOperator*{\argminB}{argmin}

\DeclareMathOperator*{\argmaxB}{argmax}

\definecolor{Gray}{gray}{0.5}
\definecolor{nicergreen}{rgb}{0.13, 0.54, 0.13}
\definecolor{nicered}{rgb}{0.83, 0.16, 0.16}
\definecolor{lightgray}{RGB}{230, 230, 230}
\definecolor{Highlight}{HTML}{39b54a}  %

\newcommand{\cgap}[2]{
\fontsize{6pt}{1em}\selectfont{\textsuperscript{${#1}${#2}}}
}
\newcommand{\cgaphl}[2]{
\fontsize{6pt}{1em}\selectfont{\textcolor{nicergreen}{\textsuperscript{${#1}$\textbf{#2}}}}
}
\newcommand{\cgaphlr}[2]{
\fontsize{6pt}{1em}\selectfont{\textcolor{nicered}{\textsuperscript{${#1}$\textbf{#2}}}}
}

\usepackage{appendix}
\usepackage{footnote}

\def\plaineqref#1{\ref{#1}}

\usepackage{microtype}
\usepackage{cuted}
\usepackage{tocloft}

\usepackage[T1]{fontenc}
\usepackage{DejaVuSans}

\usepackage{ifsym, marvosym}

\let\svthefootnote\thefootnote
\newcommand\freefootnote[1]{%
  \let\thefootnote\relax%
  \footnotetext{#1}%
  \let\thefootnote\svthefootnote%
}

%% file: sections/00_abstract.tex
\begin{abstract}
Existing dynamic data pruning methods often fail under noisy-label settings, as they typically rely on per-sample loss as the ranking criterion.
This could mistakenly lead to preserving noisy samples due to their high loss values, resulting in significant performance drop.
To address this, we propose \textbf{\algname{}}, a noise-robust module designed to enhance the reliability of dynamic pruning under label noise.
Specifically, \algname{} introduces the Dynamic Alignment Score (DAS), which is a loss-trajectory-based criterion that enables more accurate identification of noisy samples, thereby improving pruning effectiveness.
As a simple yet effective plug-and-play module, \algname{} can be seamlessly integrated into state-of-the-art dynamic pruning frameworks, consistently outperforming them without modifying either the model architecture or the training pipeline.
Extensive experiments on five widely-used benchmarks across various noise types and pruning ratios demonstrate the effectiveness of \algname{}, boosting accuracy by up to 6.3\% over state-of-the-art baselines.
Our results offer a generalizable solution for pruning under noisy data, encouraging further exploration of learning in real-world scenarios.
Code is available at: \url{https://github.com/leonqin430/AlignPrune}.

\end{abstract}

%% file: sections/01_intro.tex
\section{Introduction}
\label{sec:intro}

Modern deep learning models are increasingly trained on large-scale datasets, resulting in substantial computational demands~\cite{radford2021learning,oquab2023dinov2,fan2025scaling,achiam2023gpt,brown2020language}.
To mitigate this, \emph{data pruning}~\cite{guo2022deepcore} has emerged as an effective strategy for reducing total training cost by discarding low-utility samples, while without degrading the final model performance. 

Existing data pruning approaches fall into two main categories, the first being \textit{static coreset selection}~\cite{toneva2018empirical,paul2021deep,park2022meta,han2018co,coleman2019selection,sener2018active,sorscher2022beyond,xia2022moderate,zhang2024spanning}, which pre-selects a fixed subset of the full dataset before training begins.
However, such static approaches are often brittle in real-world scenarios.
The challenge of label noise has become increasingly crucial in the current era of large-scale models, where the immense cost of curating perfectly clean data presents a major bottleneck. 
Modern large-scale datasets are therefore often noisy~\cite{kisel2024flaws}, 
due to weak image-text supervision~\cite{radford2021learning}, automatic web crawlers~\cite{oquab2023dinov2}, or inherent semantic ambiguity~\cite{singh2024benchmarking}.
Once a noisy or uninformative sample is included in the selected subset, it remains in the training loop permanently, inevitably degrading the model's performance.

\begin{figure*}[t]
    \centering
    \includegraphics[width=\linewidth]{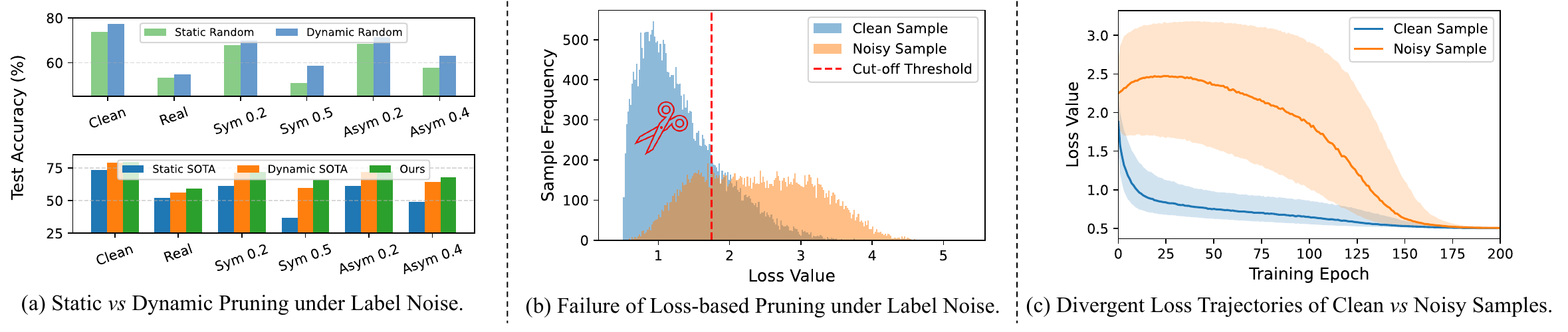}
    \caption{\textbf{Illustrations of: 
    (a) Our Motivation: 
    Performance comparison of static \vs dynamic pruning methods under label noise.}
    (\textit{upper}) Dynamic pruning is inherently more robust to noise than static pruning. Random dynamic pruning consistently outperforms random static pruning across all noise types.
    (\textit{bottom}) This motivates our work to further improve dynamic methods, where our AlignPrune achieves a new state-of-the-art, consistently outperforming dynamic (InfoBatch~\cite{qin2024infobatch}) and static-robust (Prune4ReL~\cite{park2023robust}, labeled as ‘Static SOTA') baselines. 
    %
    %
    \textbf{(b) Our Observation: Failure of loss-based pruning methods.}  
    Existing methods use a loss threshold (dashed line) that is unreliable, which could mistakenly prune many low-loss clean samples while retain high-loss noisy ones.
    As pruning progresses, the accumulation of noisy examples in the training loop degrades the overall model performance.
    \textbf{(c) Our Insight: Divergent Loss Trajectories of Clean \vs Noisy Samples.}
    We find that clean samples typically follow a smooth, monotonic decrease in loss, while noisy samples exhibit non-monotonic loss patterns, highlighting the potential of trajectory-based reference signals in separating clean from noisy data. 
    %
    }
    \vspace{-10pt}
    \label{fig:teaser}
\end{figure*}

\textit{Dynamic data pruning}~\cite{raju2021accelerating,qin2024infobatch,zhou2025scale,okanovic2024repeated} offers a more robust alternative by adaptively updating the subset selection during model training. 
By exposing the model to a broader portion of the dataset throughout training, this paradigm allows for the revision of which samples to keep or discard, making it more suitable under noisy scenarios.
As shown in~\cref{fig:teaser}\textcolor{cvprblue}{a}, dynamic methods consistently outperform static methods (even Prune4ReL~\cite{park2023robust}, a method especially designed for robust pruning) across all noise types, suggesting the inherent advantage of dynamic pruning under label noise.

Despite this promise, current state-of-the-art dynamic pruning methods, such as InfoBatch~\cite{qin2024infobatch} and SeTa~\cite{zhou2025scale}, possess a critical vulnerability: they typically rank each training sample based on their individual loss values.
This criterion becomes unreliable under label noise, as noisy samples often tend to produce large losses~\cite{jiang2018mentornet}.
Consequently, existing methods mistakenly preserve or even prioritize noisy samples during the pruning process, resulting in degraded performance under noisy-label scenarios, as illustrated in~\cref{fig:teaser}\textcolor{cvprblue}{b}.

To address this limitation, we are the first to investigate dynamic data pruning under label noise, proposing \textbf{\textit{\algname{}}}, a noise-robust and effective plug-and-play module for improving the reliability of dynamic pruning. 
The key idea is to replace the loss-based ranking criteria with a noisy-robust signal: the proposed Dynamic Alignment Score (DAS), which quantifies the correlation between each sample's loss trajectory and the average loss from a reference set over time.

The core intuition is that clean and noisy samples exhibit fundamentally different learning dynamics.
Clean samples tend to follow a consistent pattern of loss decay as the model improves, while noisy samples exhibit more erratic or inconsistent behavior, as shown in~\cref{fig:teaser}\textcolor{cvprblue}{c}.
By comparing these temporal patterns, DAS can effectively provide a robust signal for accurately filtering out noisy data samples.

\algname{} can be seamlessly integrated into existing dynamic pruning methods~\cite{qin2024infobatch,zhou2025scale}, by replacing their loss-based sample ranking with our loss-trajectory-based criterion. 
The resulting framework preserves the unbiasedness of the gradient expectation of vanilla dynamic pruning method, while improving resilience to noise without requiring any change to the model architecture or the training pipeline.

%
We evaluate \algname{} on five widely-used benchmarks: CIFAR-100N, CIFAR-10N~\cite{wei2021learning}, WebVision~\cite{li2017webvision}, Clothing-1M~\cite{xiao2015learning} and ImageNet-1K~\cite{deng2009imagenet}, under a wide range of noisy-label conditions. 
Experiments demonstrate that \algname{} consistently outperforms both static and dynamic baselines in noisy-label settings, while maintaining competitive performance under the clean-label case.
Our method also improves training efficiency, yielding higher accuracy while also reducing the total time cost.

%% file: sections/02_related.tex
\section{Related Work}
\label{sec:related work}

\paragraph{Static Data Pruning.}

Static pruning methods
~\cite{guo2022deepcore,toneva2018empirical,paul2021deep,garg2023samples,park2022meta,han2018co,coleman2019selection,he2024large,cho2025lightweight,sener2018active,sorscher2022beyond,xia2022moderate,zhang2024spanning,tan2025data,tan2023data,agarwal2020contextual,zheng2022coverage,zheng2025elfs,mirzasoleiman2020coresets,killamsetty2021grad,killamsetty2021glister,iyer2021submodular,nagaraj2025finding} 
aim to select a fixed subset of the original dataset that yields comparable performance to training on the whole dataset.
Existing approaches rely on pre-defined or heuristic metrics, which can be broadly categorized into several types: 
informative-based~\cite{tan2025data,tan2023data},
error-based~\cite{toneva2018empirical, paul2021deep}, 
geometry-based~\cite{agarwal2020contextual, sener2018active},
gradient-matching-based~\cite{mirzasoleiman2020coresets, killamsetty2021grad}, 
uncertainty-based~\cite{coleman2019selection,he2024large,cho2025lightweight}, 
coverage-based~\cite{zheng2022coverage,zheng2025elfs},  
bilevel-optimization-based~\cite{killamsetty2021glister},
memorization-based~\cite{agiollo2024approximating,garg2023samples},
sub-modularity-based methods~\cite{iyer2021submodular}.
More recently,~\cite{nagaraj2025finding} introduce the concept of using loss trajectories for static coreset selection, which is conceptually related to our approach.
However, their method is limited to clean-label and only static pruning scenarios, which also lacks a theoretically grounded justification despite achieving lossless results.

\vspace{-9pt}
\paragraph{Dynamic Data Pruning.}

Dynamic pruning methods
~\cite{raju2021accelerating,qin2024infobatch,zhou2025scale,guo2024scan,zhao2025differential,hassan2025rcap,yuan2025instance,yang2025dynamic,wu2025partial,hong2024diversified,zhao2025differential,okanovic2024repeated} 
aim to reduce total training cost by selecting a sequence of training subsets over epochs, while maintaining model performance. 
RS2~\cite{okanovic2024repeated} provides a dynamic pruning baseline that learns from randomly sampled data throughout training.
More recent methods typically perform epoch-wise pruning based on a scoring criterion.
For example,~\cite{raju2021accelerating} leverages a dynamic uncertainty estimation, \cite{qin2024infobatch,zhou2025scale,hassan2025rcap} rely directly on per-sample loss values for ranking.
IES~\cite{yuan2025instance} utilizes second-order loss dynamics to identify informative samples, and~\cite{yang2025dynamic} unifies dynamic data selection and augmentation under the dynamic pruning.
Some methods~\cite{wu2025partial,hong2024diversified} apply sample selection within each training batch.
Recently,~\cite{guo2024scan} and~\cite{zhao2025differential} extend this dynamic pruning strategy to contrastive pre-training frameworks.
Despite promising results, none of the above explore dynamic pruning under noisy-label settings. 
Current approaches remain vulnerable in such scenarios due to their reliance on non-robust ranking signals, which often lead to the repeated selection and training of noisy samples.

\vspace{-9pt}
\paragraph{Noisy Label Learning.}

Noisy Label Learning~\cite{song2022learning,li2022neighborhood,zhang2023rankmatch,lyu2019curriculum,li2020dividemix,liu2020early,liu2022robust,park2023robust} has been extensively studied to improve the robustness of deep networks in real-world settings.
Recent methods primarily adopt self-supervised or semi-supervised techniques to re-label noisy data via consistency regularization~\cite{li2020dividemix,liu2020early,liu2022robust,park2023robust}. 
For example, DivideMix~\cite{li2020dividemix} employs co-training with a dynamic mixture model, ELR+~\cite{liu2020early} introduces entropy-based regularization, and SOP+~\cite{liu2022robust} incorporates label smoothing with learnable consistency targets. 
In addition, Prune4ReL~\cite{park2023robust} proposes a noise-aware pruning method, aiming to select subsets to maximize re-labeling accuracy and downstream generalization performance.
While effective, these methods often require substantial computation overhead or multiple-model training, limiting their practicality in large-scale settings.
However, data pruning methods offer a natural solution to scalability by reducing the number of samples processed while maintaining or even improving training performance. 
Our approach builds on this idea by integrating data pruning techniques into noisy label learning.

%% file: sections/03_method.tex
\section{Methodology}
\label{sec:methodology}

In this section, we present \emph{\algname{}}, a plug-and-play dynamic data pruning module designed to improve model robustness under label noise. 

\subsection{Preliminaries}
\label{subsec:preliminary}

We focus on the task of data pruning for robust classification task under noisy label scenarios, which is a setting widely studied in the machine learning community.
Given a noisy dataset $\mathcal{D} = \{z_i\}|_{i=1}^{|\mathcal{D}|} = \{(x_i, \tilde{y}_i)\}|_{i=1}^{|\mathcal{D}|}$, where $x_i \in \mathcal{X} \subset \mathbb{R}^d$ denotes the input and $\tilde{y}_i$ denotes the corresponding label, which is possibly noisy.
The goal of data pruning is to find a smaller but most informative subset $\mathcal{S} \subset \mathcal{D}$ with $|\mathcal{S}| < |\mathcal{D}|$\footnote{For static pruning, we aim to identify the most representative subset $\mathcal{S}$ from $\mathcal{D}$ according to specific selection criterion before training.
While for dynamic pruning, we aim to identify a sequence of subsets $\mathcal{S}=\{\mathcal{S}^{(t)}\}_{t=1}^T$ across each epoch,
where $\mathcal{S}^{(t)} \subset \mathcal{D}$ with $|\mathcal{S}^{(t)}| < |\mathcal{D}|$ for epoch $t$.
},
such that a model $\theta_{\mathcal{S}}$ trained on $\mathcal{S}$ achieves comparable or better performance than the model $\theta_{\mathcal{D}}$ trained on the full dataset $\mathcal{D}$.
Formally, we aim to find the optimal subset $\mathcal{S}^{*}$ satisfying:
\begin{equation}
\begin{gathered}
\mathcal{S}^{*} = \argmaxB_{\mathcal{S}:\ |{\mathcal{S}}| \leq s} ~\sum_{(x,\tilde{y}) \in \mathcal{D}} \mathbbm{1}_ {[{f(x;{{\theta}_{\mathcal{S}}})~=~y^*}] }  :  {\theta}_{\mathcal{S}}=\argminB_{{\theta}} ~ \mathcal{L}(\theta; \mathcal{S})\,,
\end{gathered}
\label{eq:goal}
\end{equation}
where $f(x;{{\theta}_{\mathcal{S}}})\in \mathbb{R}^c$ is the model's prediction vector, $y^*$ is the ground-truth label of a noisy example $x$, and $s$ is the target subset size. 

However, finding the optimal subset $\mathcal{S}^{*}$ through direct optimization of Eq.\,\plaineqref{eq:goal} is infeasible in practice, because the ground-truth labels $y^*$ are unknown. 
Therefore, existing approaches~\cite{park2023robust,qin2024infobatch} often rely on heuristics (loss values, or certain conditions), to identify informative or clean examples. 
However, these methods break down in the presence of noisy labels.
To address this, we propose a robust, dynamic sample-ranking criterion in the next section, which estimates label correctness based on trajectory-level consistency with clean behavior. 
This enables reliable pruning without needing the correct ground-truth label $y^*$.

\begin{figure*}[t!]
    \centering
    \includegraphics[width=0.88\linewidth]{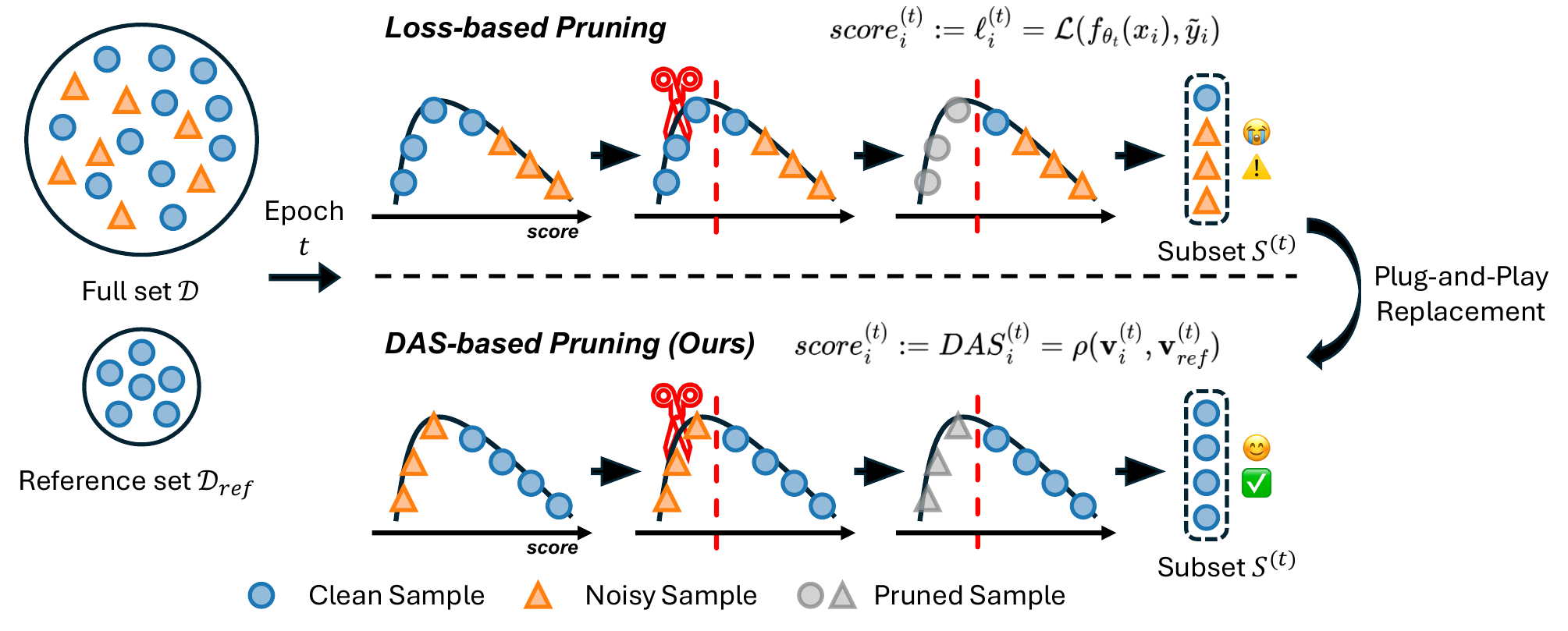}
    \vspace{-8pt}
    \caption{\textbf{Illustration of the proposed plug-and-play \textit{\algname{}} module.} 
    Existing dynamic pruning methods~\cite{qin2024infobatch,zhou2025scale} rank samples based on their loss values, which could mistakenly result in a final subset $\mathcal{S}^{(t)}$ retained with noisy samples. 
    Our method replaces the ranking score with DAS, where noisy samples tend to have low scores to be correctly pruned, resulting in a robust subset $\mathcal{S}^{(t)}$ composed primarily of clean samples.
    Best viewed in color.}
    \vspace{-10pt}
    \label{fig:framework}
\end{figure*}

\subsection{Dynamic Alignment Score}
\label{subsec:loss_traj}

Previous dynamic data pruning methods~\cite{raju2021accelerating,qin2024infobatch,zhou2025scale} typically rely on per-epoch training loss to rank the importance of each sample. 
While this is effective in clean-label settings with grounded theoretical supports in~\cite{qin2024infobatch}, it becomes unreliable with the introduction of label noise. 
Noisy samples often incur large losses and are mistakenly preserved during training~\cite{jiang2018mentornet}, leading to degraded performance. 
To address this, we propose a new ranking criterion named the Dynamic Alignment Score (DAS), which evaluates the correctness of the label of training sample by measuring the consistency of its loss trajectory with a clean reference signal.

\vspace{-6pt}
\paragraph{Loss Trajectory.}

To capture the learning dynamics of each sample, we define a sample's loss trajectory as the sequence of its training loss values over a fixed window of recent epochs.
Let $\ell_i^{(t)}$ denote the training loss of $i$-th sample at epoch $t$, computed as: 
\begin{equation}
\ell_i^{(t)} = \mathcal{L}(f_{\theta_t}(x_i), \tilde{y}_i),
\label{eq:2}
\end{equation}
where $\theta_t$ is the trained model at epoch $t$, $f_{\theta_t}(x_i)$ is model’s prediction, and $\tilde{y}_i$ is the (potentially noisy) label.
Given a trajectory window size of $N$ epochs, the loss trajectory vector $\mathbf{v}^{(t)}_i \in \mathbb{R}^T$ for the $i$-th sample at epoch $t$ is defined as:
\begin{equation}
\mathbf{v}^{(t)}_i = \left[\ell_i^{(t - N + 1)}, \ell_i^{(t - N + 2)}, \ldots, \ell_i^{(t)}\right].
\label{eq:3}
\end{equation}

To obtain a clean reference, we utilize a small held-out validation set $\mathcal{D}_{ref}$, which is assumed to be free of label noise\footnote{We treat this as an idealized assumption and will empirically evaluate both the size and purity of this reference set in~\cref{subsubsec:dependence_on_clean_data}.}. 
For each epoch $t'$, we compute the average reference loss and the resulting reference loss trajectory $\mathbf{v}^{(t')}_{ref} \in \mathbb{R}^T$:
\begin{align}
\bar{\ell}_{ref}^{(t')} = \frac{1}{|\mathcal{D}_{ref}|} \sum_{(x_j, \tilde{y_j})\in\mathcal{D}_{ref}} \mathcal{L}(f_{\theta_{t'}}(x_j), \tilde{y}_j) ~~ ,
\label{eq:4}
\end{align}
\vspace{-6pt}
\begin{align}
\mathbf{v}^{(t')}_{ref} = \left[\bar\ell_{ref}^{(t' - N + 1)}, \bar\ell_{ref}^{(t' - N + 2)}, \ldots, \bar\ell_{ref}^{(t')}\right].
\label{eq:5}
\end{align}

\paragraph{Correlation-Based Alignment.}

With the loss trajectories of training samples $\mathbf{v}_i$ and the clean validation reference $\mathbf{v}_{ref}$ defined, we now introduce the \textit{Dynamic Alignment Score} (DAS), which provides a correlation-based metric that quantifies how closely each training sample aligns with the overall trend of clean reference performance. 
The key intuition is that samples with correct labels $\tilde{y} =y^*$ typically follow a consistent trajectory of decreasing loss that mirrors the behavior of a clean reference set, while noisy samples exhibit more erratic or inconsistent patterns. 
The proposed DAS quantifies this alignment by computing the correlation score between the training sample’s loss trajectory and the reference trajectory.
While hard clean samples may also exhibit high loss, their learning dynamics still differ significantly\footnote{More detailed discussions on hard \vs noisy samples are in~\cref{subsec:additional_analysis_hard_vs_noisy}.} from truly noisy samples.

Formally, given a training sample’s loss trajectory $\mathbf{v}^{(t)}_i$ and the clean validation reference $\mathbf{v}^{(t)}_{ref}$ at epoch $t$, we define the Dynamic Alignment Score as:
\begin{equation}
    {DAS}^{(t)}_i = \rho(\mathbf{v}^{(t)}_i, \mathbf{v}^{(t)}_{ref}),
    \label{eq:6}
\end{equation}
where $\rho$ denotes the correlation function.
In our implementation, Pearson’s correlation~\cite{pearson1895vii} is used due to its scale invariance and computational efficiency.

A positive DAS value indicates synchronized learning dynamics with the expected clean loss pattern, suggesting the training sample is likely to be clean. 
A negative DAS value, on the other hand, reflects a conflicting learning dynamics with clean reference behavior, implying the presence of label noise.
In practice, DAS is computed in batch using vectorized matrix operations and updated at the end of each epoch.
The per-sample trajectories are stored in a memory bank of window size $N$, allowing for efficient computation without excessive overhead. 
While we use Pearson correlation as the default $\rho$, other similarity metrics will be discussed further in the ablation studies in~\cref{sec:experiments}.

\begin{algorithm}[htbp]
\caption{\algname{} for Dynamic Data Pruning}
\label{alg:loss_traj}
    \begin{algorithmic}[1]
    {\small 
        \REQUIRE Noisy training set $\mathcal{D}$, Clean reference set $\mathcal{D}_{ref}$, Dynamic data pruning method $\mathcal{M}$ (with pruning probability $r$~\hyperlink{alg:fn1}{\textsuperscript{a}}), Total epochs $T$, Trajectory window size $N$, Correlation function $\rho$
        
        \STATE Initialize model parameters $\theta^{(0)}$
        
        \STATE \textbf{for} epoch $t=1,...,T$ \textbf{do}
        \STATE ~~ \textbf{for} each sample $z_i \in \mathcal{S}^{(t)} \subset \mathcal{D}$ \textbf{do}
        \STATE ~~ ~~ Compute training loss $\ell_i^{(t)}$ by Eq.~\plaineqref{eq:2}
        \STATE ~~ ~~ Update trajectory $\mathbf{v}^{(t)}_i$ with $\ell_i^{(t)}$ by Eq.~\plaineqref{eq:3}
        \STATE ~~ \textbf{end for}
        \STATE ~~ Compute average reference loss $\bar\ell_{ref}^{(t)}$ over $\mathcal{D}_{ref}$ by Eq.~\plaineqref{eq:4}
        \STATE ~~ Update reference trajectory $\mathbf{v}^{(t)}_{ref}$ with $\bar\ell_{ref}^{(t)}$ by Eq.~\plaineqref{eq:5}
        \STATE ~~ Compute $DAS_i^{(t)}=\rho(\mathbf{v}^{(t)}_i, \mathbf{v}^{(t)}_{ref})$ for each $z_i \in \mathcal{S}^{(t)}$
        \STATE ~~ Update $\theta^{(t)}$ on $\mathcal{S}^{(t)}$ following the update rule of $\mathcal{M}$~~\hyperlink{alg:fn2}{\textsuperscript{b}}
        \STATE ~~ Apply pruning: $\mathcal{S}^{(t+1)} \leftarrow \mathcal{M}(\mathcal{S}^{(t)}, DAS, r)$ 
        \STATE \textbf{end for}

        \ENSURE Final trained model $\theta^{(T)}$
    }
    \end{algorithmic}
    \vspace{-9pt}
    \noindent\rule{0.4\columnwidth}{0.4pt}
    \par\footnotesize
    \hypertarget{alg:fn1}{\textsuperscript{a}} Pruning probability $r$ is a hyper-parameter used in dynamic pruning methods; see details in \cref{subsec:experimental_setup}.\\
    \hypertarget{alg:fn2}{\textsuperscript{b}} The specific gradient update rule is determined by the base pruning method $\mathcal{M}$. For instance, InfoBatch uses an expectation rescaling operation to ensure unbiased gradients, which is preserved when using \algname{}.
\end{algorithm}

\subsection{\algname{}: A Noise-Robust Dynamic Pruning Module}
\label{subsec:alignprune}

Building on the \textit{Dynamic Alignment Score} (DAS), we now describe how to integrate it into existing dynamic data pruning frameworks. 
As shown in~\cref{fig:framework}, our proposed module, \emph{\algname{}}, serves as a plug-and-play replacement for loss-based ranking in dynamic methods
such as InfoBatch~\cite{qin2024infobatch} and SeTa~\cite{zhou2025scale}, enabling improved robustness under noisy label scenarios.

In prior dynamic pruning methods designed for clean labels, each training sample is ranked based on its per-epoch loss.
The pruning is then performed either by discarding samples below the mean loss~\cite{qin2024infobatch}, or via a sliding-window strategy to remove ineffective samples over time~\cite{zhou2025scale}. 
However, as discussed in~\cref{subsec:loss_traj}, loss values alone are unreliable indicators under label noise. 
To address this, \emph{\algname{}} replaces the loss-based ranking metric with the proposed DAS:
\vspace{-3pt}
\begin{equation}
    {score}^{(t)}_i := {DAS}^{(t)}_i,
\end{equation}
where a higher score indicates stronger alignment with clean reference behavior, and thus a higher likelihood of the sample being clean.
Training on subsets with higher DAS values enables the model to focus on more reliable samples and mitigates the negative impact of noisy labels.
The algorithm of the plug-and-play \emph{\algname{}} is outlined in~\Cref{alg:loss_traj}.

Importantly, \emph{\algname{}} modifies only the sample ranking strategy while preserving the base model architecture, original training loop, and previous pruning mechanism,
ensuring seamless integration to existing dynamic pruning frameworks.
In~\cref{sec:experiments}, we demonstrate that this simple modification consistently improves performance across multiple datasets and various noisy label conditions, while maintaining training efficiency.

%% file: sections/04_experiment.tex
\section{Experiments}
\label{sec:experiments}

\input{tab/results_cifar100}
\input{tab/results_cifar10}

\subsection{Experimental Setup}\label{subsec:experimental_setup}

\paragraph{Datasets.}

We evaluate \emph{\algname{}} on four benchmark datasets commonly used in noisy-label research: \textbf{CIFAR-100N}, \textbf{CIFAR-10N}~\cite{wei2021learning}, \textbf{WebVision}~\cite{li2017webvision} and \textbf{Clothing-1M}~\cite{xiao2015learning}.
CIFAR-N datasets facilitate controlled studies under both \textit{synthetic} and \textit{real} noise conditions, while WebVision and Clothing-1M offer large-scale settings with real-world noise.
Additionally, we include \textbf{ImageNet-1K}~\cite{deng2009imagenet} with \textit{synthetic} noise for further large-scale experiments.

\vspace{-9pt}
\paragraph{Baseline Methods.}

We compare \textit{\algname{}} with static and dynamic pruning methods, along with their respective random selection baselines~\cite{okanovic2024repeated}.\footnote{Due to space constraint, comprehensive comparison including both static and dynamic pruning methods is in~\cref{tab:supp_results_cifar100_full,tab:supp_results_cifar10_full} in~\cref{subsec:extended_baseline_comparisons}.}
For static pruning, we consider representative methods including: SmallL~\cite{jiang2018mentornet}, Margin~\cite{coleman2019selection}, Forgetting~\cite{toneva2018empirical}, Glister~\cite{killamsetty2021glister}, GraNd~\cite{paul2021deep}, Moderate~\cite{xia2022moderate}, SSP~\cite{sorscher2022beyond}, Prune4ReL~\cite{park2023robust}, Dyn-Unc~\cite{he2024large} and DUAL~\cite{cho2025lightweight}.
For dynamic pruning, we benchmark two methods: InfoBatch~\cite{qin2024infobatch} and SeTa~\cite{zhou2025scale}.
Additional discussions on the baseline selection are provided in~\cref{subsec:implementation_details}.

\vspace{-12pt}
\paragraph{Implementation Details.}
For each of the data pruning baselines, we adopt the default optimal hyper-parameter settings to ensure fairness and reproducibility.
As for our proposed \emph{\algname{}}, we align the pruning probability $r$\footnote{To avoid confusion, we denote the pruning probability $r$ as the fraction of low-score samples discarded in each epoch, and the pruning ratio / prune ratio as the target percentage of the entire dataset that is pruned over the training process, both following the terminology in~\cite{qin2024infobatch}.} with the value used in the underlying dynamic pruning methods~\cite{qin2024infobatch,zhou2025scale}.
$N$ is set to 25 by default, and reduced to 5 for Clothing-1M due to its shorter fine-tuning duration.
Pearson's correlation is used as the default $\rho$.
For dataset with an official validation split, we use this official split as the clean reference set.
For dataset with only train/test splits, we randomly sample a subset from the training split (with equal size to the test split) as the clean reference set.
All experiments are conducted on NVIDIA RTX A6000 GPUs. 
Additional training and reproducibility details are provided in~\cref{subsec:implementation_details}.

\vspace{-9pt}
\paragraph{Evaluation.}
Following prior work~\cite{qin2024infobatch,zhou2025scale}, we evaluate each method at pruning ratios of $\{$30\%, 50\%, 70\%$\}$ for all the datasets.
For dynamic pruning methods, the pruning ratio is controlled by adjusting the number of training epochs to match the total number of forward passes in the full-training baseline, as in~\cite{qin2024infobatch}. 
Each experiment is \textit{repeated three times}, and we report the average accuracy from the final epoch.

\subsection{Main Results}

\paragraph{Performance Comparisons.}

\Cref{tab:results_cifar100} and~\Cref{tab:results_cifar10} present the test accuracy of baseline methods and \algname{} on CIFAR-100N and CIFAR-10N under three pruning ratios.
Overall, \algname{} consistently achieves the best performance across all settings, demonstrating strong robustness under various types of label noise.
Numerically, on CIFAR-100N, InfoBatch with \algname{} can achieve an average accuracy improvement of {+2.7\%} under {30\%} pruning ratio across all noise types, compared to only {+0.6\%} improvement from vanilla InfoBatch.
On CIFAR-10N, SeTa with \algname{} can also yield an average {+0.1\%} improvement under {50\%} pruning ratio, whereas vanilla SeTa shows a significant drop of {-1.4\%} in the same setting.
These results demonstrate that \algname{} can be effectively integrated into dynamic methods to improve performance under both real and synthetic label noise, without sacrificing the effectiveness when dealing with clean data.
See~\cref{tab:supp_results_statistical_analysis} for statistical analysis.

\begin{figure}[t]
    \centering
      {\includegraphics[width=\linewidth]{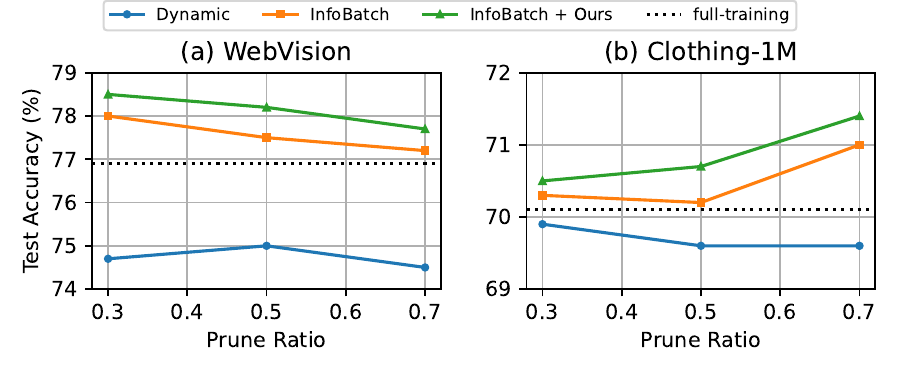}}
      \vspace{-22pt}
\caption{\textbf{Data pruning performance comparison: (a) WebVision; (b) Clothing-1M.} Best viewed in color.}
\vspace{-18pt}
\label{fig:results_comparison}
\end{figure}

\input{tab/results_imagenet}

\begin{figure*}[htbp]
    \centering
    \begin{subfigure}[t]{0.325\textwidth}
        \includegraphics[width=\linewidth]{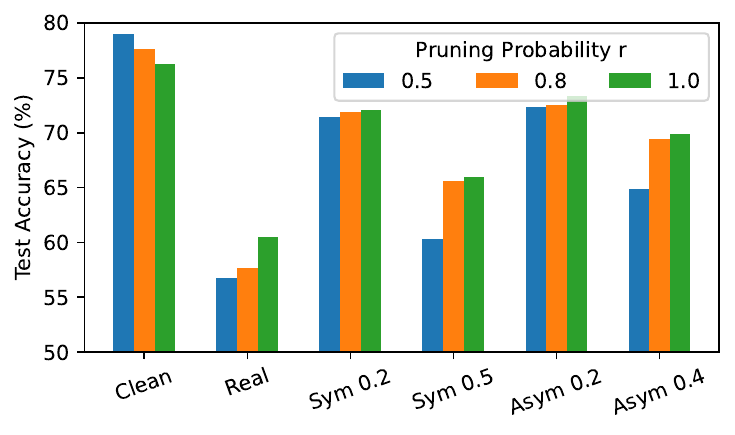}
        \caption{Ablation on $r$.}
        \label{fig:ablation_r}
    \end{subfigure}
    \begin{subfigure}[t]{0.325\textwidth}
        \includegraphics[width=\linewidth]{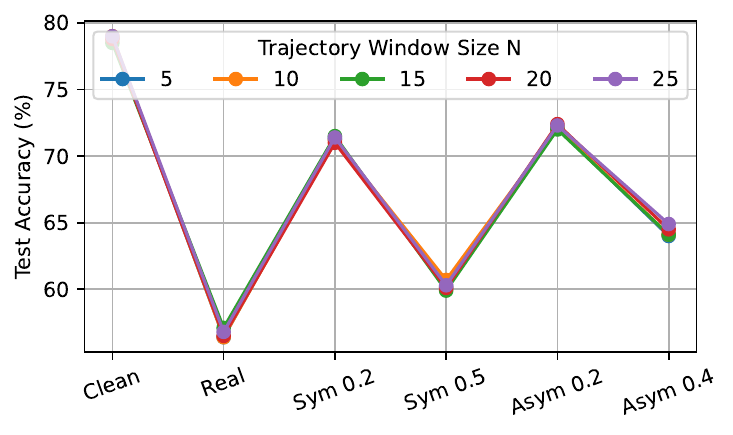}
        \caption{Ablation on $N$.}
        \label{fig:ablation_N}
    \end{subfigure}
    \begin{subfigure}[t]{0.325\textwidth}
        \includegraphics[width=\linewidth]{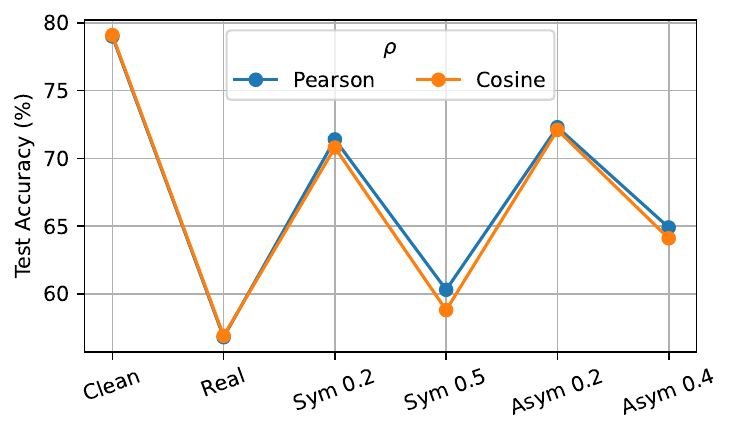}
        \caption{Ablation on $\rho$.}
        \label{fig:ablation_rho}
    \end{subfigure}
    \vspace{-5pt}
    \caption{\textbf{Ablation on hyper-parameter selection on CIFAR-100N.}
    Best viewed in color.}
    \label{fig:ablations}
    \vspace{-12pt}
\end{figure*}
%

To further validate our approach, ~\Cref{fig:results_comparison} presents a comparison between dynamic pruning baselines and \algname{} on two large-scale datasets: WebVision and Clothing-1M.
Due to instability, the training of SeTa collapses on these datasets, hence we report results only for InfoBatch and its \algname{} variant.
In line with the observations on CIFAR-N datasets, \algname{} outperforms existing baselines across different pruning ratios.
Quantitatively, InfoBatch with \algname{} achieves an average improvement of {+0.5\%} over vanilla InfoBatch, even under the fine-tuning duration of Clothing-1M.
These findings reinforce the effectiveness of \algname{} in dynamic pruning, showcasing its scalability and robustness across large-scale datasets in both from-scratch training and fine-tuning settings.

Furthermore, \algname{} can be seamlessly integrated with existing robust learning techniques to achieve state-of-the-art performance.
Full results are detailed in~\cref{tab:supp_results_cifar100_with_sop,tab:supp_results_cifar10_with_sop} in~\cref{subsec:extended_baseline_comparisons}.

\vspace{-15pt}
\paragraph{Efficiency Comparisons.}
In addition to the performance comparison, we compare the training time cost between baseline methods and \algname{}.
Specifically, we use the default setting to train CIFAR-100N with a pruning ratio of {50\%} under the Real noisy type.
As reported in~\cref{tab:efficiency_comparison}, under identical computational conditions, our method not only improves accuracy but also enhances training efficiency.
While the DAS calculation introduces a negligible per-epoch overhead (see~\cref{tab:supp_prune_ratio_comparison} in~\cref{subsec:additional_analysis_efficiency} for a detailed analysis), 
the resulting improvement in sample selection efficiency leads to a significant reduction in the total training time compared to both the baseline methods and full-data training.

\input{tab/overhead_experiment}

\subsection{Large-scale Experimental Results}

To further validate the scalability and robustness of \algname{}, we extend our experiments to ImageNet-1K~\cite{deng2009imagenet} classification under both clean and noisy-label settings, across modern CNN-based~\cite{liu2022convnet} and ViT-based~\cite{touvron2021training,liu2021swin} architectures beyond ResNets.
As shown in~\Cref{tab:results_imagenet}, \algname{} consistently improves performance across various settings, confirming its ability to scale effectively to larger datasets and architectures.

\input{tab/results_pseudo_val_ablation}

\subsection{Ablation Studies}

\subsubsection{Effect of Hyper-Parameters}
We conduct ablation experiments to investigate the effect of key hyper-parameters in \algname{}, using CIFAR-100N with default settings across five noisy-label types and the clean-label case.

\vspace{-9pt}
\paragraph{Effect of Pruning Probability $r$.}

We vary the pruning probability $r$ from 0.5 to 1.0 and evaluate performance under all six label conditions. 
As shown in~\cref{fig:ablation_r}, increasing $r$ leads to decreased accuracy when trained on clean labels, while improves accuracy under all noisy-label types.
This highlights that the proposed \algname{} effectively identifies and discards noisy samples with a higher pruning probability, which also suggests that a lower $r$ should be used under clean-label case.

\vspace{-9pt}
\paragraph{Effect of Trajectory Window Size $N$.}

We evaluate window sizes $N$ ranging from 5 to 25, to examine the sensitivity of the loss trajectory length.
As shown in~\cref{fig:ablation_N}, \algname{} performs consistently across the range of $N$, indicating low sensitivity to the trajectory window size. 
In practice, we use $N=25$, which corresponds to $12.5\%$ of the default 200 training epochs.
We provide more detailed results on this ablation in~\cref{tab:supp_ablations_window_size} in~\cref{subsec:additional_analysis_window_size}.

\vspace{-6pt}
\paragraph{Effect of Correlation Function $\rho$.}

We compare two similarity functions used to compute the Dynamic Alignment Score: Pearson correlation and cosine similarity.
As shown in~\cref{fig:ablation_rho}, both yield similar performance across all noise types. 
We adopt Pearson correlation by default due to its scale invariance and computational efficiency.
We provide further discussions on this ablation in~\cref{subsec:additional_analysis_correlation}.

\subsubsection{Dependence on Clean Data}
\label{subsubsec:dependence_on_clean_data}

To evaluate the robustness of \algname{} in realistic scenarios where obtaining a large clean reference set is challenging, we conduct two additional ablation studies below with CIFAR-100N. 

\vspace{-6pt}
\paragraph{Performance with Limited Clean Data.}

We first assess the sensitivity of \algname{} to the size of the clean reference set by varying its fraction from $100\%$ to $0.1\%$.
This setup tests the method's ability to operate with minimal clean supervision.

\begin{figure}[t]
    \centering
    \vspace{2pt}
    {\includegraphics[width=\linewidth]{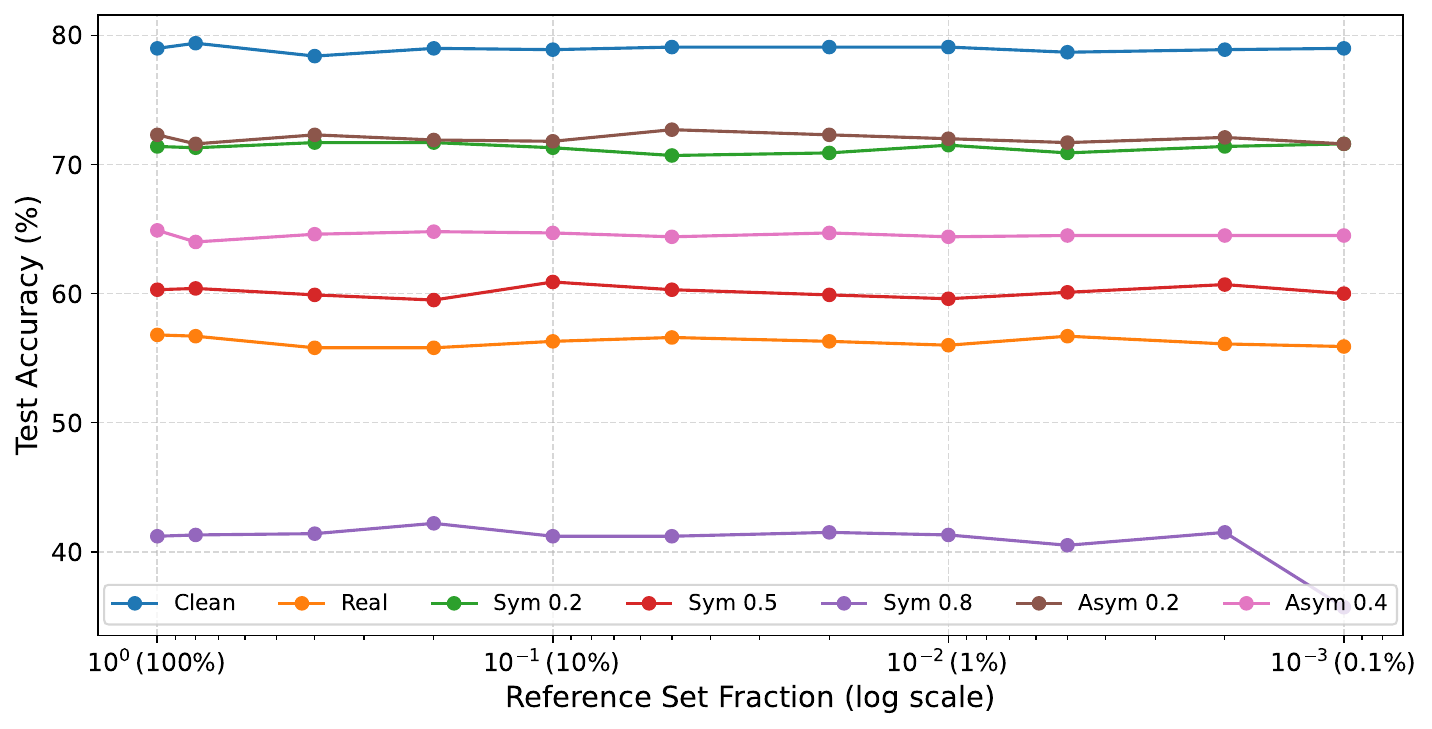}}
    \vspace{-15pt}
    \caption{\textbf{Ablation on \algname{} with varying reference set sizes.}
    Reference set size is reduced from $100\%$ to $0.1\%$ of the original one.
    \algname{} remains effective even with minimal clean supervision.
    Best viewed in color.}
    \vspace{-15pt}
\label{fig:ablation_smaller_val}
\end{figure}

As shown in~\cref{fig:ablation_smaller_val}, \algname{} maintains consistent performance across all six noisy-label types, as well as the clean-label condition, even with significantly reduced reference data scale.
Remarkably, with only 10 clean samples ($0.1\%$ of the original clean set), the performance remains comparable to the full reference setting.  
A slight performance drop is observed for symmetric noise with noise rate of 0.8 under the $0.1\%$ reference scale.
We attribute this to insufficient clean reference signal under extreme noisy condition.
However, increasing the clean set fraction to just $0.2\%$ (doubling to 20 clean samples) can effectively mitigate this drop, showing the robustness of \algname{} even under high-noisy cases.

\vspace{-6pt}
\paragraph{Performance with Pseudo Clean Data.}

We further evaluate \algname{}’s dependency on having a clean reference set by replacing it with estimated pseudo-clean subsets selected directly from the noisy training data.
Since most prior works do not aim to isolate purely clean samples, we use subsets generated by SmallL~\cite{jiang2018mentornet} and Moderate~\cite{xia2022moderate}, two static coreset methods known for identifying subsets with low noise levels (as reported in Table \textcolor{cvprblue}{4} of~\cite{park2023robust}).

\cref{tab:results_pseudo_val_ablation} shows that using these pseudo-clean subsets yields comparable performance to using real clean reference sets, across both clean and noisy label conditions, even when the pseudo-clean subset is reduced to just $1\%$ of the reference set scale.
We also use a random subset from the original noisy training set as the reference for comparison.
Results show that while this performs similarly in the clean-label setting (where train and reference distributions match), it leads to significant accuracy drops under noisy conditions.
This ablation confirms that \algname{} can remain effective even when no explicit clean data is available, provided that a reasonably clean subset can be estimated \textbf{using noisy or even unlabeled data}, thus further highlighting its practical flexibility and robustness.
We provide further discussions on the intuitions behind above ablation studies in~\cref{subsec:additional_analysis_dependence_clean_data}.

%


%% file: tab/results_cifar100.tex
\begin{table*}[ht]
    \centering
    \footnotesize
    \setlength{\tabcolsep}{3pt}
    \caption{
    \textbf{Classification results on CIFAR-100N with ResNet-18.} 
    Performance gaps relative to the full-training setting are indicated as superscripts. 
    The mean $\Delta$ is computed across all noise types.}
    \vspace{-3pt}
    \begin{tabular}{lcccccccc}
        \toprule
        Noisy Type $\rightarrow$ & \multirow{2}{*}{Clean} & \multirow{2}{*}{Real} & \multicolumn{3}{c}{Symmetric} & \multicolumn{2}{c}{Asymmetric} & Mean \\ 
        \cmidrule(lr){4-6}
        \cmidrule(lr){7-8}
        Pruning Method $\downarrow$ & & & 0.2 & 0.5 & 0.8 & 0.2 & 0.4 & $\Delta$ \\
        \midrule
        Full-training 
        & 78.2 & 56.1 & 71.4 & 58.6 & 39.8 & 72.4 & 63.3 & -- \\
        \midrule
        \multicolumn{9}{c}{\textbf{Prune Ratio $\sim$30\%}} \\
        \midrule
        Static Random 
        & 73.8\cgaphlr{-}{4.4} & 53.3\cgaphlr{-}{2.8} & 67.9\cgaphlr{-}{3.4} & 51.1\cgaphlr{-}{7.5} & 30.3\cgaphlr{-}{9.5} & 68.4\cgaphlr{-}{4.0} & 57.8\cgaphlr{-}{5.5} & -5.3 \\
        Dynamic Random~\cite{okanovic2024repeated}
        & 77.3\cgaphlr{-}{0.9} & 54.7\cgaphlr{-}{1.4} & 69.9\cgaphlr{-}{1.4} & 58.8\cgaphl{+}{0.2} & 40.1\cgaphl{+}{0.3} & 71.5\cgaphlr{-}{0.9} & 63.2\cgaphlr{-}{0.1} & -0.6 \\
        InfoBatch~\citep{qin2024infobatch}
        & 79.0\cgaphl{+}{0.8} & 56.1\cgap{+}{0.0} & 71.4\cgap{+}{0.0} & 59.7\cgaphl{+}{1.1} & 41.8\cgaphl{+}{2.0} & 71.9\cgaphlr{-}{0.5} & 64.2\cgaphl{+}{0.9} & +0.6 \\
        \textbf{InfoBatch + Ours} 
        & \cellcolor{cyan!20}79.3\cgaphl{+}{1.1} & \cellcolor{cyan!20}59.4\cgaphl{+}{3.3} & \cellcolor{cyan!20}71.8\cgaphl{+}{0.4} & \cellcolor{cyan!20}66.0\cgaphl{+}{7.4} & \cellcolor{cyan!20}41.8\cgaphl{+}{2.0} & \cellcolor{cyan!20}72.6\cgaphl{+}{0.2} & \cellcolor{cyan!20}68.0\cgaphl{+}{4.7} & \cellcolor{cyan!20}\textbf{+2.7} \\
        SeTa~\citep{zhou2025scale} 
        & 79.0\cgaphl{+}{0.8} & 55.6\cgaphlr{-}{0.5} & 70.2\cgaphlr{-}{1.2} & 59.0\cgaphl{+}{0.4} & 41.6\cgaphl{+}{1.8} & 71.4\cgaphlr{-}{0.9} & 63.2\cgaphlr{-}{0.1} & +0.0 \\
        \textbf{SeTa + Ours}  
        & \cellcolor{cyan!20}79.3\cgaphl{+}{1.1} & \cellcolor{cyan!20}56.3\cgaphl{+}{0.2} & \cellcolor{cyan!20}70.8\cgaphlr{-}{0.5} & \cellcolor{cyan!20}60.5\cgaphl{+}{1.9} & \cellcolor{cyan!20}41.6\cgaphl{+}{1.8} & \cellcolor{cyan!20}71.9\cgaphlr{-}{0.5} & \cellcolor{cyan!20}64.3\cgaphl{+}{1.0} & \cellcolor{cyan!20}\textbf{+0.7} \\
        \midrule
        \multicolumn{9}{c}{\textbf{Prune Ratio $\sim$50\%}} \\
        \midrule
        Static Random 
        & 72.1\cgaphlr{-}{6.1} & 51.8\cgaphlr{-}{4.3} & 64.0\cgaphlr{-}{7.4} & 47.2\cgaphlr{-}{11.5} & 22.9\cgaphlr{-}{16.9} & 65.3\cgaphlr{-}{7.0} & 54.6\cgaphlr{-}{8.7} & -8.8 \\
        Dynamic Random~\cite{okanovic2024repeated}  
        & 75.3\cgaphlr{-}{2.9} & 54.1\cgaphlr{-}{2.0} & 70.4\cgaphlr{-}{1.0} & 59.5\cgaphl{+}{0.9} & 40.7\cgaphl{+}{0.9} & 70.1\cgaphlr{-}{2.3} & 64.8\cgaphl{+}{1.6} & -0.7 \\
        InfoBatch~\citep{qin2024infobatch}
        & 77.7\cgaphlr{-}{0.5} & 56.0\cgaphlr{-}{0.1} & 71.3\cgaphlr{-}{0.1} & 60.5\cgaphl{+}{1.9} & 42.2\cgaphl{+}{2.4} & 71.8\cgaphlr{-}{0.6} & 65.2\cgaphl{+}{2.0} & +0.7 \\
        \textbf{InfoBatch + Ours}  
        & \cellcolor{cyan!20}78.5\cgaphl{+}{0.3} & \cellcolor{cyan!20}60.7\cgaphl{+}{4.6} & \cellcolor{cyan!20}71.6\cgaphl{+}{0.3} & \cellcolor{cyan!20}62.0\cgaphl{+}{3.4} & \cellcolor{cyan!20}42.6\cgaphl{+}{2.8} & \cellcolor{cyan!20}72.6\cgaphl{+}{0.3} & \cellcolor{cyan!20}68.6\cgaphl{+}{5.3} & \cellcolor{cyan!20}\textbf{+2.4} \\
        SeTa~\citep{zhou2025scale} 
        & 77.5\cgaphlr{-}{0.7} & 55.7\cgaphlr{-}{0.4} & 70.7\cgaphlr{-}{0.7} & 60.0\cgaphl{+}{1.4} & 40.5\cgaphl{+}{0.7} & 71.9\cgaphlr{-}{0.5} & 64.5\cgaphl{+}{1.2} & +0.2 \\
        \textbf{SeTa + Ours} 
        & \cellcolor{cyan!20}78.4\cgaphl{+}{0.2} & \cellcolor{cyan!20}56.3\cgaphl{+}{0.1} & \cellcolor{cyan!20}71.2\cgaphlr{-}{0.2} & \cellcolor{cyan!20}61.0\cgaphl{+}{2.4} & \cellcolor{cyan!20}41.9\cgaphl{+}{2.1} & \cellcolor{cyan!20}72.2\cgaphlr{-}{0.2} & \cellcolor{cyan!20}66.0\cgaphl{+}{2.7} & \cellcolor{cyan!20}\textbf{+1.0} \\
        \midrule
        \multicolumn{9}{c}{\textbf{Prune Ratio $\sim$70\%}} \\
        \midrule
        Static Random 
        & 69.7\cgaphlr{-}{8.5} & 45.9\cgaphlr{-}{10.2} & 56.7\cgaphlr{-}{14.6} & 39.0\cgaphlr{-}{19.6} & 15.8\cgaphlr{-}{24.0} & 58.4\cgaphlr{-}{13.9} & 49.5\cgaphlr{-}{13.8} & -14.9 \\
        Dynamic Random~\cite{okanovic2024repeated}  
        & 75.2\cgaphlr{-}{3.0} & 53.3\cgaphlr{-}{2.9} & 70.2\cgaphlr{-}{1.2} & 62.7\cgaphl{+}{4.1} & 41.0\cgaphl{+}{1.2} & 71.9\cgaphlr{-}{0.5} & 67.3\cgaphl{+}{4.0} & +0.3 \\
        InfoBatch~\citep{qin2024infobatch} 
        & 77.1\cgaphlr{-}{1.1} & 55.0\cgaphlr{-}{1.2} & 71.4\cgaphl{+}{0.1} & 63.6\cgaphl{+}{5.0} & 42.4\cgaphl{+}{2.6} & 72.2\cgaphlr{-}{0.2} & 67.8\cgaphl{+}{4.5} & +1.4 \\
        \textbf{InfoBatch + Ours} 
        & \cellcolor{cyan!20}77.5\cgaphlr{-}{0.7} & \cellcolor{cyan!20}58.3\cgaphl{+}{2.2} & \cellcolor{cyan!20}72.2\cgaphl{+}{0.9} & \cellcolor{cyan!20}64.7\cgaphl{+}{6.1} & \cellcolor{cyan!20}42.8\cgaphl{+}{3.0} & \cellcolor{cyan!20}72.5\cgaphl{+}{0.1} & \cellcolor{cyan!20}69.1\cgaphl{+}{5.9} & \cellcolor{cyan!20}\textbf{+2.5} \\
        SeTa~\citep{zhou2025scale}
        & 77.2\cgaphlr{-}{1.1} & 55.2\cgaphlr{-}{1.0} & 71.6\cgaphl{+}{0.2} & 62.4\cgaphl{+}{3.8} & 42.4\cgaphl{+}{2.6} & 72.0\cgaphlr{-}{0.4} & 67.4\cgaphl{+}{4.1} & +1.2 \\
        \textbf{SeTa + Ours} 
        & \cellcolor{cyan!20}77.8\cgaphlr{-}{0.4} & \cellcolor{cyan!20}55.5\cgaphlr{-}{0.6} & \cellcolor{cyan!20}72.3\cgaphl{+}{0.9} & \cellcolor{cyan!20}63.3\cgaphl{+}{4.7} & \cellcolor{cyan!20}42.7\cgaphl{+}{2.9} & \cellcolor{cyan!20}72.7\cgaphl{+}{0.3} & \cellcolor{cyan!20}67.6\cgaphl{+}{4.3} & \cellcolor{cyan!20}\textbf{+1.7} \\
        \bottomrule
    \end{tabular}
    \vspace{-6pt}
    \label{tab:results_cifar100}
\end{table*}

%% file: tab/results_cifar10.tex
\begin{table*}[t]
    \centering
    \footnotesize
    \setlength{\tabcolsep}{3pt}
    \caption{\textbf{Classification results on CIFAR-10N with ResNet-18.} 
    Performance gaps relative to the full-training setting are indicated as superscripts. 
    The mean $\Delta$ is computed across all noise types.}
    \vspace{-3pt}
    \begin{tabular}{lccccccccc}
        \toprule
        Noisy Type $\rightarrow$ & \multirow{2}{*}{Clean} & \multicolumn{2}{c}{Real} & \multicolumn{3}{c}{Symmetric} & \multicolumn{2}{c}{Asymmetric} & Mean \\ 
        \cmidrule(lr){3-4}
        \cmidrule(lr){5-7}
        \cmidrule(lr){8-9}
        Pruning Method $\downarrow$ & & Real-A & Real-W & 0.2 & 0.5 & 0.8 & 0.2 & 0.4 & $\Delta$ \\
        \midrule
        Full-training 
        & 95.6 & 90.7 & 78.3 & 91.3 & 85.4 & 65.6 & 90.6 & 85.8 & -- \\
        \midrule
        \multicolumn{10}{c}{\textbf{Prune Ratio $\sim$30\%}} \\
        \midrule
        Static Random 
        & 94.6\cgaphlr{-}{1.0} & 89.9\cgaphlr{-}{0.7} & 75.0\cgaphlr{-}{3.3} & 89.0\cgaphlr{-}{2.3} & 81.2\cgaphlr{-}{4.2} & 57.6\cgaphlr{-}{8.0} & 87.9\cgaphlr{-}{2.8} & 82.5\cgaphlr{-}{3.3} & -3.2 \\
        Dynamic Random~\cite{okanovic2024repeated}  
        & 94.8\cgaphlr{-}{0.8} & 90.3\cgaphlr{-}{0.3} & 77.3\cgaphlr{-}{1.0} & 90.8\cgaphlr{-}{0.5} & 84.2\cgaphlr{-}{1.2} & 64.8\cgaphlr{-}{0.8} & 91.5\cgaphl{+}{0.9} & 84.5\cgaphlr{-}{1.3} & -0.6 \\
        InfoBatch~\citep{qin2024infobatch} 
        & 95.3\cgaphlr{-}{0.3} & 91.0\cgaphl{+}{0.3} & 78.1\cgaphlr{-}{0.3} & 92.0\cgaphl{+}{0.7} & 85.6\cgaphl{+}{0.1} & 67.5\cgaphl{+}{1.9} & 91.3\cgaphl{+}{0.7} & 87.0\cgaphl{+}{1.3} & +0.5 \\
        \textbf{InfoBatch + Ours} 
        & \cellcolor{cyan!20}95.3\cgaphlr{-}{0.3} & \cellcolor{cyan!20}91.4\cgaphl{+}{0.7} & \cellcolor{cyan!20}82.0\cgaphl{+}{3.7} & \cellcolor{cyan!20}92.7\cgaphl{+}{1.4} & \cellcolor{cyan!20}87.9\cgaphl{+}{2.5} & \cellcolor{cyan!20}67.8\cgaphl{+}{2.2} & \cellcolor{cyan!20}92.9\cgaphl{+}{2.3} & \cellcolor{cyan!20}90.0\cgaphl{+}{4.2} & \cellcolor{cyan!20}\textbf{+2.1} \\
        SeTa~\citep{zhou2025scale}
        & 95.2\cgaphlr{-}{0.4} & 90.6\cgaphlr{-}{0.1} & 76.3\cgaphlr{-}{2.0} & 91.8\cgaphl{+}{0.5} & 84.7\cgaphlr{-}{0.7} & 64.2\cgaphlr{-}{1.5} & 90.8\cgaphl{+}{0.1} & 88.4\cgaphl{+}{2.6} & -0.2 \\
        \textbf{SeTa + Ours}  
        & \cellcolor{cyan!20}95.3\cgaphlr{-}{0.3} & \cellcolor{cyan!20}90.7\cgap{+}{0.0} & \cellcolor{cyan!20}79.1\cgaphl{+}{0.8} & \cellcolor{cyan!20}92.5\cgaphl{+}{1.2} & \cellcolor{cyan!20}85.0\cgaphlr{-}{0.5} & \cellcolor{cyan!20}64.6\cgaphlr{-}{1.0} & \cellcolor{cyan!20}91.8\cgaphl{+}{1.2} & \cellcolor{cyan!20}89.3\cgaphl{+}{3.5} & \cellcolor{cyan!20}\textbf{+0.6} \\
        \midrule
        \multicolumn{10}{c}{\textbf{Prune Ratio $\sim$50\%}} \\
        \midrule
        Static Random 
        & 93.3\cgaphlr{-}{2.3} & 88.6\cgaphlr{-}{2.1} & 72.4\cgaphlr{-}{5.9} & 87.4\cgaphlr{-}{3.9} & 77.5\cgaphlr{-}{8.0} & 51.4\cgaphlr{-}{14.2} & 85.8\cgaphlr{-}{4.9} & 77.9\cgaphlr{-}{7.9} & -6.1 \\
        Dynamic Random~\cite{okanovic2024repeated}  
        & 94.5\cgaphlr{-}{1.1} & 89.2\cgaphlr{-}{1.5} & 76.1\cgaphlr{-}{2.3} & 88.9\cgaphlr{-}{2.4} & 83.7\cgaphlr{-}{1.8} & 63.3\cgaphlr{-}{2.3} & 90.6\cgap{+}{0.0} & 83.7\cgaphlr{-}{2.1} & -1.7 \\
        InfoBatch~\citep{qin2024infobatch}
        & 95.0\cgaphlr{-}{0.6} & 90.5\cgaphlr{-}{0.2} & 77.7\cgaphlr{-}{0.6} & 92.0\cgaphl{+}{0.7} & 87.5\cgaphl{+}{2.1} & 64.3\cgaphlr{-}{1.3} & 92.0\cgaphl{+}{1.3} & 87.5\cgaphl{+}{1.7} & +0.4 \\
        \textbf{InfoBatch + Ours}  
        & \cellcolor{cyan!20}95.0\cgaphlr{-}{0.6} & \cellcolor{cyan!20}91.3\cgaphl{+}{0.6} & \cellcolor{cyan!20}82.4\cgaphl{+}{4.1} & \cellcolor{cyan!20}92.8\cgaphl{+}{1.5} & \cellcolor{cyan!20}87.6\cgaphl{+}{2.1} & \cellcolor{cyan!20}64.5\cgaphlr{-}{1.1} & \cellcolor{cyan!20}92.1\cgaphl{+}{1.5} & \cellcolor{cyan!20}89.8\cgaphl{+}{4.0} & \cellcolor{cyan!20}\textbf{+1.5} \\
        SeTa~\citep{zhou2025scale}
        & 94.8\cgaphlr{-}{0.8} & 89.4\cgaphlr{-}{1.3} & 78.5\cgaphl{+}{0.2} & 91.3\cgap{+}{0.0} & 86.0\cgaphl{+}{0.5} & 53.0\cgaphlr{-}{12.7} & 91.7\cgaphl{+}{1.1} & 87.9\cgaphl{+}{2.1} & -1.4 \\
        \textbf{SeTa + Ours} 
        & \cellcolor{cyan!20}94.9\cgaphlr{-}{0.7} & \cellcolor{cyan!20}89.7\cgaphlr{-}{1.0} & \cellcolor{cyan!20}79.3\cgaphl{+}{1.0} & \cellcolor{cyan!20}91.9\cgaphl{+}{0.6} & \cellcolor{cyan!20}87.6\cgaphl{+}{2.2} & \cellcolor{cyan!20}60.2\cgaphlr{-}{5.5} & \cellcolor{cyan!20}92.2\cgaphl{+}{1.6} & \cellcolor{cyan!20}88.5\cgaphl{+}{2.7} & \cellcolor{cyan!20}\textbf{+0.1} \\
        \midrule
        \multicolumn{10}{c}{\textbf{Prune Ratio $\sim$70\%}} \\
        \midrule
        Static Random 
        & 90.2\cgaphlr{-}{5.4} & 86.6\cgaphlr{-}{4.1} & 69.0\cgaphlr{-}{9.3} & 85.3\cgaphlr{-}{6.0} & 73.8\cgaphlr{-}{11.6} & 41.4\cgaphlr{-}{24.3} & 82.3\cgaphlr{-}{8.3} & 75.3\cgaphlr{-}{10.5} & -9.9 \\
        Dynamic Random~\cite{okanovic2024repeated}  
        & 93.0\cgaphlr{-}{2.6} & 87.4\cgaphlr{-}{3.2} & 75.9\cgaphlr{-}{2.5} & 87.4\cgaphlr{-}{3.9} & 84.6\cgaphlr{-}{0.8} & 60.2\cgaphlr{-}{5.5} & 91.4\cgaphl{+}{0.8} & 86.8\cgaphl{+}{1.0} & -2.1 \\
        InfoBatch~\citep{qin2024infobatch} 
        & 94.3\cgaphlr{-}{1.3} & 89.9\cgaphlr{-}{0.8} & 79.4\cgaphl{+}{1.0} & 92.4\cgaphl{+}{1.1} & 87.0\cgaphl{+}{1.6} & 61.0\cgaphlr{-}{4.7} & 92.5\cgaphl{+}{1.8} & 88.3\cgaphl{+}{2.5} & +0.2 \\
        \textbf{InfoBatch + Ours} 
        & \cellcolor{cyan!20}94.3\cgaphlr{-}{1.3} & \cellcolor{cyan!20}90.8\cgaphl{+}{0.1} & \cellcolor{cyan!20}81.0\cgaphl{+}{2.7} & \cellcolor{cyan!20}92.8\cgaphl{+}{1.5} & \cellcolor{cyan!20}87.9\cgaphl{+}{2.5} & \cellcolor{cyan!20}61.1\cgaphlr{-}{4.5} & \cellcolor{cyan!20}92.9\cgaphl{+}{2.3} & \cellcolor{cyan!20}89.2\cgaphl{+}{3.4} & \cellcolor{cyan!20}\textbf{+0.8} \\
        SeTa~\citep{zhou2025scale} 
        & 94.3\cgaphlr{-}{1.3} & 89.5\cgaphlr{-}{1.2} & 77.7\cgaphlr{-}{0.6} & 92.2\cgaphl{+}{0.9} & 87.1\cgaphl{+}{1.7} & 30.4\cgaphlr{-}{35.2} & 92.2\cgaphl{+}{1.5} & 88.1\cgaphl{+}{2.3} & -4.0 \\
        \textbf{SeTa + Ours} 
        & \cellcolor{cyan!20}94.5\cgaphlr{-}{1.1} & \cellcolor{cyan!20}90.0\cgaphlr{-}{0.7} & \cellcolor{cyan!20}79.4\cgaphl{+}{1.1} & \cellcolor{cyan!20}92.6\cgaphl{+}{1.3} & \cellcolor{cyan!20}87.7\cgaphl{+}{2.3} & \cellcolor{cyan!20}58.0\cgaphlr{-}{7.6} & \cellcolor{cyan!20}92.5\cgaphl{+}{1.9} & \cellcolor{cyan!20}88.5\cgaphl{+}{2.7} & \cellcolor{cyan!20}\textbf{+0.0} \\
        \bottomrule
    \end{tabular}
    \vspace{-6pt}
    \label{tab:results_cifar10}
\end{table*}

%% file: tab/results_imagenet.tex
\begin{table*}[t]
    \centering
    \footnotesize
    \setlength{\tabcolsep}{3pt}
    \vspace{-5pt}
    \caption{\textbf{Classification results on ImageNet-1K across modern CNN and ViT architectures.} 
    Performance gaps relative to the full-training setting are indicated as superscripts.}
    \begin{tabular}{p{6pt}lccccccc}
    \toprule
    & Model Architecture $\rightarrow$ & \multicolumn{2}{c}{ConvNeXt~\cite{liu2022convnet}} & \multicolumn{2}{c}{DeiT~\cite{touvron2021training}} & \multicolumn{2}{c}{Swin~\cite{liu2021swin}} & Mean \\ 
        \cmidrule(lr){3-4}
        \cmidrule(lr){5-6}
        \cmidrule(lr){7-8}    
    & Pruning Method $\downarrow$ & Tiny & Base & Small & Base & Tiny & Base & $\Delta$ \\
    \midrule
    \multirow{3}{*}{\rotatebox{90}{Clean}} & Full-training & 73.3 & 75.6 & 71.9 & 77.8 & 73.5 & 76.9 & -- \\
    & InfoBatch & 73.2\cgaphlr{-}{0.1} & 75.4\cgaphlr{-}{0.2} & 71.5\cgaphlr{-}{0.4} & \textbf{77.8}\cgaphl{+}{0.0} & \textbf{73.6}\cgaphl{+}{0.1} & 77.1\cgaphl{+}{0.2} & -0.1 \\
    & \textbf{InfoBatch + Ours} 
    & \cellcolor{cyan!20}\textbf{73.4}\cgaphl{+}{0.1} 
    & \cellcolor{cyan!20}\textbf{75.7}\cgaphl{+}{0.1} 
    & \cellcolor{cyan!20}\textbf{72.1}\cgaphl{+}{0.2} 
    & \cellcolor{cyan!20}\textbf{77.8}\cgaphl{+}{0.0} 
    & \cellcolor{cyan!20}\textbf{73.6}\cgaphl{+}{0.1} 
    & \cellcolor{cyan!20}\textbf{77.2}\cgaphl{+}{0.3} 
    & \cellcolor{cyan!20}\textbf{+0.1} \\
    
    \midrule
    \multirow{3}{*}{\rotatebox{90}{Noisy}} & Full-training & 66.4 & 67.3 & 64.7 & 68.9 & 65.0 & 67.8 & -- \\
    & InfoBatch & 65.8\cgaphlr{-}{0.6} & 66.8\cgaphlr{-}{0.5} & 64.1\cgaphlr{-}{0.6} & 68.1\cgaphlr{-}{0.8} & 64.3\cgaphlr{-}{0.7} & 66.9\cgaphlr{-}{0.9} & -0.7 \\
    & \textbf{InfoBatch + Ours} 
    & \cellcolor{cyan!20}\textbf{67.2}\cgaphl{+}{0.8} 
    & \cellcolor{cyan!20}\textbf{67.8}\cgaphl{+}{0.5} 
    & \cellcolor{cyan!20}\textbf{65.5}\cgaphl{+}{0.8} 
    & \cellcolor{cyan!20}\textbf{69.4}\cgaphl{+}{0.5} 
    & \cellcolor{cyan!20}\textbf{65.6}\cgaphl{+}{0.6} 
    & \cellcolor{cyan!20}\textbf{68.2}\cgaphl{+}{0.4}
    & \cellcolor{cyan!20}\textbf{+0.6} \\
    
    \bottomrule
    \end{tabular}
    \label{tab:results_imagenet}
\end{table*}

%% file: tab/overhead_experiment.tex
\begin{table}[t]
    \centering
    \footnotesize
    \vspace{2pt}
    \setlength{\tabcolsep}{1.8pt}
    \caption{\textbf{Efficiency comparison on CIFAR-100N.} Results are reported with ResNet-18 under 50\% prune ratio with Real noise for 200 epochs on two RTX-A6000-GPUs. 
    ``Ours'' refers to AlignPrune with InfoBatch.}
    \vspace{-6pt}
    \begin{tabular}{lccccc|c}
        \toprule
        & Static & Dynamic & SeTa & InfoBatch & \textbf{Ours} & Full Data \\
        \midrule
        Acc (\%) $\uparrow$ & 51.8 & 54.1 & 55.7 & 56.0 & \textbf{60.7} & 56.1 \\
        Time (mins) $\downarrow$ & 24.7 & 24.4 & 25.8 & 23.9 & \textbf{22.5} & 46.7 \\
        \bottomrule
    \end{tabular} 
    \vspace{-15pt}
    \label{tab:efficiency_comparison}
\end{table}

%% file: tab/results_pseudo_val_ablation.tex
\begin{table*}[t]
    \centering
    \footnotesize
    \setlength{\tabcolsep}{5pt}
    \caption{
    \textbf{Ablation on \algname{} with estimated pseudo clean set.} 
    We evaluate \algname{} using pseudo clean subsets selected by SmallL~\citep{jiang2018mentornet} and Moderate~\citep{xia2022moderate} as  reference sets. 
    A random subset from noisy training set is also included for comparison.}
    \vspace{-3pt}
    \begin{tabular}{lcccccc}
        \toprule
        Reference-set Fraction $\rightarrow$ & \multicolumn{2}{c}{$100\%$} & \multicolumn{2}{c}{$10\%$} & \multicolumn{2}{c}{$1\%$} \\ 
        \cmidrule(lr){2-3}
        \cmidrule(lr){4-5}
        \cmidrule(lr){6-7}
        Reference-set Type $\downarrow$ & Clean & Real & Clean & Real & Clean & Real \\
        \midrule
        \rowcolor{lightgray}
        Noisy Train-set
        & 78.9 & 53.7 & 79.0 & 54.1 & 79.0 & 53.8 \\
        Clean Validation-set
        & 79.0 & 56.8 & 78.9 & 56.3 & 79.1 & 56.0 \\
        Pseudo Clean-set (estimated by~\cite{jiang2018mentornet})
        & 79.1 & 56.0 & 79.1 & 56.8 & 79.0 & 56.4 \\
        Pseudo Clean-set (estimated by~\cite{xia2022moderate})
        & 78.6 & 56.5 & 79.2 & 56.4 & 78.7 & 55.9 \\
        \bottomrule
    \end{tabular}
    \vspace{-9pt}
    \label{tab:results_pseudo_val_ablation}
\end{table*}

%% file: sections/10_conclusion.tex
\section{Conclusion}
\label{sec:conclusion}
\vspace{-3pt}
In this paper, we address the critical issue of label noise in dynamic data pruning, where existing loss-based methods often fail by mistakenly preserving high-loss noisy samples than the clean ones. 
To this end, we propose \textbf{\algname{}}, a noise-robust and effective plug-and-play module that introduces a more reliable sample selection criterion: Dynamic Alignment Score (DAS). 
By leveraging the correlation between sample loss trajectories and a clean reference signal, \algname{} effectively enables more accurate identification and pruning of noisy samples, which naturally generalizes to various label noise conditions.
Extensive experiments on five benchmarks confirm the effectiveness of this proposed approach consistently across diverse real-world noisy-label conditions. 
Our results offer a generalizable solution for pruning under noisy
data, encouraging future exploration of learning in real-world scenarios for broader applicability, especially in the era of large-scale models.

%% file: sections/12_appendix.tex
\maketitlesupplementary
\appendix

\cftsetindents{section}{0em}{2.1em}
\cftsetindents{subsection}{2.1em}{2.9em}
\cftsetindents{subsubsection}{5.0em}{3.8em}
\setlength{\cftbeforesecskip}{6pt}
\setlength{\cftbeforesubsecskip}{2pt}
\setlength{\cftbeforesubsubsecskip}{2pt}

\setcounter{section}{0}
\setcounter{table}{0}
\setcounter{figure}{0}
\setcounter{equation}{0}

\pagenumbering{gobble}
\begin{strip}
    \vspace{-1cm} 
    \hypersetup{linkcolor=cvprblue}
    \large \tableofcontents
    \vspace{3pt} 
\end{strip}
\clearpage
\pagenumbering{arabic}
\setcounter{page}{13}


\renewcommand{\thesection}{\Alph{section}}
\renewcommand{\thetable}{\Alph{table}}
\renewcommand{\thefigure}{\Alph{figure}}
\renewcommand{\theequation}{\Alph{equation}}

\section{Additional Implementation Details}
\label{subsec:implementation_details}

We provide further implementation details to support reproducibility of our experimental setup below.

\subsection{Details of Dataset}

Specifically, \textbf{CIFAR-100N} and \textbf{CIFAR-10N}~\cite{wei2021learning} consist of 50K human re-annotated training images of size 32 $\times$ 32, with 100 and 10 classes respectively, adapted from the original CIFAR-100 and CIFAR-10 datasets~\cite{krizhevsky2009learning}.
Both datasets include human-annotated noisy labels.
We use the Real noisy-label set from CIFAR-100N, and both the Aggregate and Worst noisy-label sets from CIFAR-10N to represent \textit{real} label noise.
Following prior work~\cite{cao2020coresets}, we inject \textit{synthetic} symmetric and asymmetric label noise with rates \{0.2, 0.5, 0.8\} and \{0.2, 0.4\} respectively, to simulate controllable noise.
\textbf{WebVision}~\cite{li2017webvision} contains 2.4M images crawled from the Web, covering the same 1,000 categories as ImageNet-1K~\cite{deng2009imagenet}.
Following standard practice in~\cite{chen2019understanding}, we use a subset of WebVision containing approximately 66K training images from the first 50 classes of the Google image.
\textbf{Clothing-1M}~\cite{xiao2015learning} consists of 1M training images sourced from online shopping platforms, with naturally occurring noisy labels.
We follow standard practice and perform fine-tuning on the entire Clothing-1M training set using a pre-trained model weights on ImageNet.
\textbf{ImageNet-1K}~\cite{deng2009imagenet} consists of 1.2M training images with 1000 classes.
We inject \textit{synthetic} asymmetric label noise with rate 0.2 following settings in~\cite{park2023robust}.

\subsection{Details of Experimental Setup}
\label{sec:implement_detail}

Table \ref{table:supp_experiment_detail} summarizes the experimental setups and hyper-parameters used across four main datasets and methods.
For CIFAR-100N and CIFAR-10N, we follow the settings from~\cite{qin2024infobatch}, using a ResNet-18~\cite{he2016deep} backbone trained for 200 epochs optimized by LARS~\cite{you2019large} with momentum of 0.9, weight decay of 5e-4, and batch size of 128. 
The initial learning rates are set to 5.2 (CIFAR-100N) and 5.62 (CIFAR-10N), both with a OneCycle learning rate scheduler (cosine annealing).
For WebVision, we adopt the standard setup from~\cite{chen2019understanding,park2023robust}, using an InceptionResNetV2~\cite{szegedy2017inception} trained for 100 epochs optimized by SGD with batch size of 32.
For Clothing-1M, we fine-tune a ResNet-50~\cite{he2016deep} pre-trained on ImageNet-1K~\cite{deng2009imagenet} for 10 epochs using a batch size of 32.
The initial learning rates of WebVision and Clothing-1M are set to 0.02 and 0.002, decayed by a factor of 10 halfway through the total training epochs.
For ImageNet-1K, we use the default settings for corresponding architectures with 100 epoch training.
Unless otherwise stated, we apply standard data augmentations, including normalization, random cropping, and horizontal flipping for all images during model training.

To ensure fairness and reproducibility, we adopt the default optimal hyper-parameter settings from the original papers for both static and dynamic data pruning baselines.
For InfoBatch~\cite{qin2024infobatch}, we use the pruning probability $r=0.5$ for CIFAR-100N and CIFAR-10N, and a more aggressive $r=0.75$ for the large-scale datasets WebVision and Clothing-1M.
For SeTa~\cite{zhou2025scale}, we use $r=0.1$ with window scale $\alpha=0.9$ and group number $k=5$ on CIFAR-100N and CIFAR-10N.
Due to training collapse of SeTa on WebVision and Clothing-1M, we only report results with InfoBatch on these two datasets.
The annealing ratio $\delta$ is set to 0.875 for both InfoBatch and SeTa across all settings.
For our \textbf{\textit{\algname{}}}, we follow the same pruning hyper-parameters as the corresponding dynamic pruning method.
The trajectory window size $N$ is set to 25 by default, and reduced to 5 for Clothing-1M due to its short fine-tuning schedule.
Pearson's correlation is used as the default correlation function $\rho$.

\subsection{Details of Noise Injection}
For \textit{synthetic} label noise on CIFAR-100N and CIFAR-10N, we follow the protocol in~\cite{cao2020coresets}.
Specifically, we inject:
\textbf{Symmetric noise} with rates \{0.2, 0.5, 0.8\}, where $r\%$ of the labels from each class $c$ are randomly flipped to the next consecutive class $c+1$, and labels from the final class are wrapped around to class $0$.
\textbf{Asymmetric noise} with rates \{0.2, 0.4\}, where selected samples are flipped within their super-class as defined in~\cite{wei2021learning}.
For ImageNet-1K experiments, we follow~\cite{park2023robust} to inject asymmetric noise with rate of 0.2.
This controlled noise injection allows us to benchmark the robustness of different baselines across varying noisy types and noisy rates.

\input{tab_supp/supp_experiment_detail}

\subsection{Details of Baseline Method}

We clarify the selection criteria for dynamic pruning baselines as follows:
Several recent methods are excluded from our evaluation due to incompatibility with our experimental setting.
Specifically, SCAN~\cite{guo2024scan} and DISSect~\cite{zhao2025differential} are tailored for contrastive pre-training frameworks like CLIP~\cite{radford2021learning}, which significantly differ from our focus on noisy learning.
Additionally, methods by~\cite{raju2021accelerating},~\cite{yang2025dynamic},~\cite{wu2025partial} and~\cite{hassan2025rcap} are excluded due to the lack of publicly available implementations, and their reported results are limited to clean-label scenarios with different hyper-parameters, making fair and reproducible comparison infeasible.

\vspace{-3pt}
\paragraph{Remark.}

Although AlignPrune is designed as a plug-and-play replacement to enhance dynamic pruning methods, its integration assumes an epoch-level sample-wise scoring and ranking mechanism based on loss trajectories.
We selected InfoBatch~\cite{qin2024infobatch} and SeTa~\cite{zhou2025scale} as representative dynamic pruning baselines because both provide public implementations that align well with our framework. 
In contrast, DivBS~\cite{hong2024diversified} and IES~\cite{yuan2025instance} adopt fundamentally different designs:
DivBS performs pruning at batch level, making per-sample loss trajectory inapplicable.
Moreover, applying such batch-level pruning under label noise tends to remove meaningful samples while retaining noisy ones.
IES, while structurally compatible, becomes functionally equivalent to InfoBatch once AlignPrune is applied, with the only difference being the use of a fixed threshold versus a dynamic mean-based threshold.
Therefore, we integrate AlignPrune only with InfoBatch and SeTa, while including DivBS and IES in the comparison for completeness.

\section{Additional Experimental Results}

\subsection{Extended Baseline Comparisons}
\label{subsec:extended_baseline_comparisons}

\input{tab_supp/supp_results_cifar100}
\input{tab_supp/supp_results_cifar10}

\paragraph{Main Results under Label Noise.} 
\Cref{tab:supp_results_cifar100_full,tab:supp_results_cifar10_full} present a comprehensive comparison of \algname{} with both static and dynamic data pruning baselines on CIFAR-100N and CIFAR-10N under three pruning ratios.
Overall, static pruning methods perform significantly worse than dynamic methods, particularly in the presence of label noise. 
This highlights the limitations of static coreset selection when noisy labels are involved.
In contrast, the proposed \algname{} consistently achieves the best performance across all pruning ratios and noise types, demonstrating strong robustness and generalizability under diverse noisy-label conditions.

\paragraph{Results with Re-labeling under Label Noise.} 

We further explore the integration of \algname{} with robust learning techniques designed to mitigate label noise. 
As highlighted in~\cite{park2023robust}, re-labeling strategies can improve model performance by explicitly correcting noisy annotations.
We adopt SOP+~\cite{liu2022robust}, a recent method designed to model label noise by disentangling noisy and clean labels through a trainable over-parameterized consistency module and self-regularization, into our framework.
Specifically, we use the original hyper-parameter settings from SOP+, and implement it on top of InfoBatch~\cite{qin2024infobatch} and Prune4ReL~\cite{park2023robust} to represent state-of-the-art dynamic and static pruning approaches, respectively.
As shown in~\cref{tab:supp_results_cifar100_with_sop,tab:supp_results_cifar10_with_sop}, SOP+ improves all methods, and when combined with \algname{}, it achieves the state-of-the-art performance across benchmarks. 
These results confirm that \algname{} remains compatible with existing re-labeling techniques and can further benefit from them under noisy scenarios.

\input{tab_supp/supp_results_cifar100_with_SOP}
\input{tab_supp/supp_results_cifar10_with_SOP}

\subsection{Additional Analysis and Discussions}\label{subsec:additional_analysis}

\subsubsection{Analysis on Efficiency Comparison}
\label{subsec:additional_analysis_efficiency}
As reported in Tab.~\ref{tab:efficiency_comparison}, \algname{} with InfoBatch can achieve better classification accuracy while reducing the total training time compared to vanilla InfoBatch.
Although \algname{} introduces a minor computational overhead due to the computation of the Dynamic Alignment Score (DAS) at each epoch, the benefit arises from its improved sample selection efficiency.
Unlike InfoBatch, which ranks samples solely based on per-epoch loss, \algname{} leverages DAS for more robust ranking, enabling more effective identification and pruning of low-utility and noisy samples. 

\begin{table}[t]
    \centering
    \footnotesize
    \vspace{2pt}
    \setlength{\tabcolsep}{4pt}
    \caption{\textbf{Average ratio of pruned samples per epoch.} Results are reported with ResNet-18 under same prune probability.
    }
    \begin{tabular}{lcc}
        \toprule
        & InfoBatch & \textbf{Ours} \\
        \midrule
        Avg. pruning ratio per epoch & 28.77\% & \textbf{32.94\%} \\
        \bottomrule
    \end{tabular} 
    \vspace{-9pt}
    \label{tab:supp_prune_ratio_comparison}
\end{table}

\input{tab_supp/supp_ablations_window_size}

Specifically, as shown in~\cref{tab:supp_prune_ratio_comparison}, replacing the original loss-based metric with DAS, and combining with the soft-limit nature of the pruning probability, leads \algname{} to achieve a higher average pruning ratio per epoch.
We argue this is not simply a side effect of more aggressive pruning, but rather a direct consequence of the superior sample selection quality of DAS. 
Because DAS provides a clearer and more confident signal for identifying noisy and uninformative samples compared to raw loss, it can decisively discard a larger fraction of the dataset at each epoch without harming performance. 
Thus, this improved selection efficiency is what drives both the accuracy gains and the reduction in total training time.

\subsubsection{Analysis on Trajectory Window Size}
\label{subsec:additional_analysis_window_size}
We provide more results on the effect of trajectory window size $N$ below.
We expand the ablation study in Fig.~\ref{fig:ablation_N} to include smaller (2, 3, 4) and larger window sizes (30, 40, 50) as shown in~\Cref{tab:supp_ablations_window_size}.
Results show that the proposed \algname{} performs consistently with window size ranging from 4 to 50 across all noisy-label types, with minimal variance presented.
However, when using extremely small window size 2 or 3, the performance show significant drop, where the loss trajectory under this condition cannot fully capture the learning dynamics as expected.
This confirms that the necessity of such trajectory window design, but also shows that \algname{} remains robust across a practical and appropriate range of values. 

\subsubsection{Analysis on Correlation Function}
\label{subsec:additional_analysis_correlation}

\begin{figure}[!t]
    \centering
      {\includegraphics[width=\linewidth]{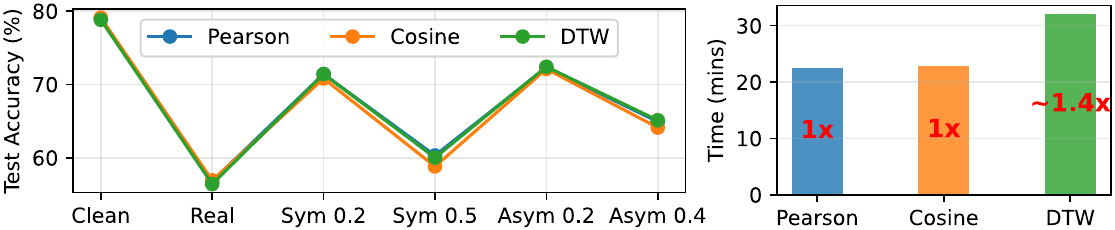}}
      \vspace{-15pt}
\caption{\textbf{Ablation on $\rho$: Accuracy (Left); Time Cost (Right).}}
\vspace{-12pt}
\label{fig:additional_ablation_rho}
\end{figure}

We provide more results on the choice of correlation function $\rho$ below.
We expand the ablation study in Fig.~\ref{fig:ablation_rho} by additionally evaluating Dynamic Time Warping (DTW) as the correlation function.
As illustrated in~\Cref{fig:additional_ablation_rho}, while DTW achieves comparable accuracy to both Pearson and Cosine similarity across all noise types, a substantially higher computational time cost can be observed.
We therefore keep to use Pearson by default to maintain optimal training efficiency without sacrificing robustness.

\subsubsection{Analysis on Fairness under Extra Reference Supervision}

\begin{figure}[!t]
    \centering
      {\includegraphics[width=\linewidth]{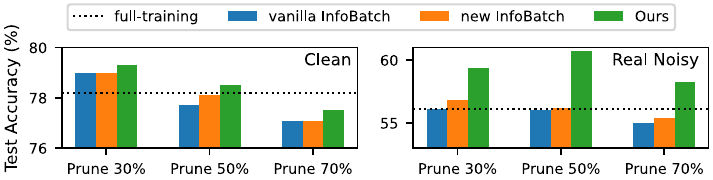}}
      \vspace{-15pt}
\caption{\textbf{Comparison between calibrated InfoBatch and Ours.}}
\vspace{-12pt}
\label{fig:fig_new_infobatch_ablation}
\end{figure}

To ensure the performance gains of AlignPrune do not stem solely from access to the extra reference data, we calibrate the loss threshold of InfoBatch using the same reference supervision. 
Specifically, we define a sample remaining region via epoch-wise reference loss statistics (using the mean and standard deviations). 
As shown in~\Cref{fig:fig_new_infobatch_ablation}, this calibrated strategy yields negligible improvement over vanilla InfoBatch, indicating the bottleneck lies in the loss-based ranking metric itself, not the access to the clean reference set.

\subsubsection{Analysis on Dependence of Clean Data}
\label{subsec:additional_analysis_dependence_clean_data}
We provide further discussions on the intuitions behind the effectiveness of \algname{} with extremely limited or even without clean references.

\vspace{-9pt}
\paragraph{Discussion on Reference Set Scale.}
From an intuitive perspective, DAS measures the degree of alignment between a sample's loss trajectory and the principal learning dynamics of the target task captured by the clean reference set. 
Even when the clean reference set is small, its averaged trajectory remains a stable indicator of the model's intended learning trend, because correlation used during DAS computation focuses on \textit{relative trends} rather than absolute loss magnitudes, which is robust to scale differences.
To provide evidence from results, 
\begin{itemize}
    \item Results in Fig.~\ref{fig:ablation_smaller_val} show remarkable stability: with only 0.1\% clean fraction, DAS provides enough signal to prune noisy samples while preserving informative ones, except under extreme noise (symmetric noise with 0.8 rate), where the clean signal becomes too weak. 
    \item Table~\ref{tab:results_pseudo_val_ablation} provides a comparison with \textit{Clean Validation-set} versus \textit{Noisy Train-set} as the reference dataset. 
    Performance remains consistent for clean-label setting but degrades significantly for noisy-label case when using noisy reference data.
    This confirms that clean reference data is crucial for providing the accurate principal learning dynamics.
\end{itemize}
In summary, these results indicate that DAS does not depend heavily on the quantity of clean data, but rather on its ability to provide a reliable signal of the principal learning dynamics.

\vspace{-9pt}
\paragraph{Discussion on Reference Set Quality.}

Using a pseudo-clean set as reference remains effective because a coreset selection method can \textit{filter} the noisy dataset to produce a subset whose average learning dynamic is statistically closer to the true clean dynamic, which can be reflected from the downstream performance of the selected coreset.
Even if this reference set is not perfectly purely clean, it has a lower noise level compared to the original noisy set, making its average trajectory a more reliable signal.
To provide evidence from results,
\begin{itemize}
    \item Table~\ref{tab:results_pseudo_val_ablation} shows that using high-quality pseudo-clean subsets (estimated by SmallL or Moderate) yields performance nearly identical to using a real clean reference set, demonstrating the practical viability of this approach in scenarios when clean data is unavailable.
    \item Table~\ref{tab:results_pseudo_val_ablation} also provides a comparison with \textit{Noisy Train-set} versus \textit{Pseudo Clean-set}.
    Performance degrades significantly under noisy scenarios, which further confirms that the quality and correctness of the reference set, not just its availability, are critical.
\end{itemize}
In summary, these results demonstrate that \algname{}'s effectiveness is not contingent on a purely clean reference set. 
Instead, it highlights the method's practical flexibility: it can remain effective provided with a coreset method capable of extracting a pseudo-clean subset that is reasonably pure to offer a reliable reference trajectory.

\begin{figure*}[t]
    \centering
    \includegraphics[width=\linewidth]{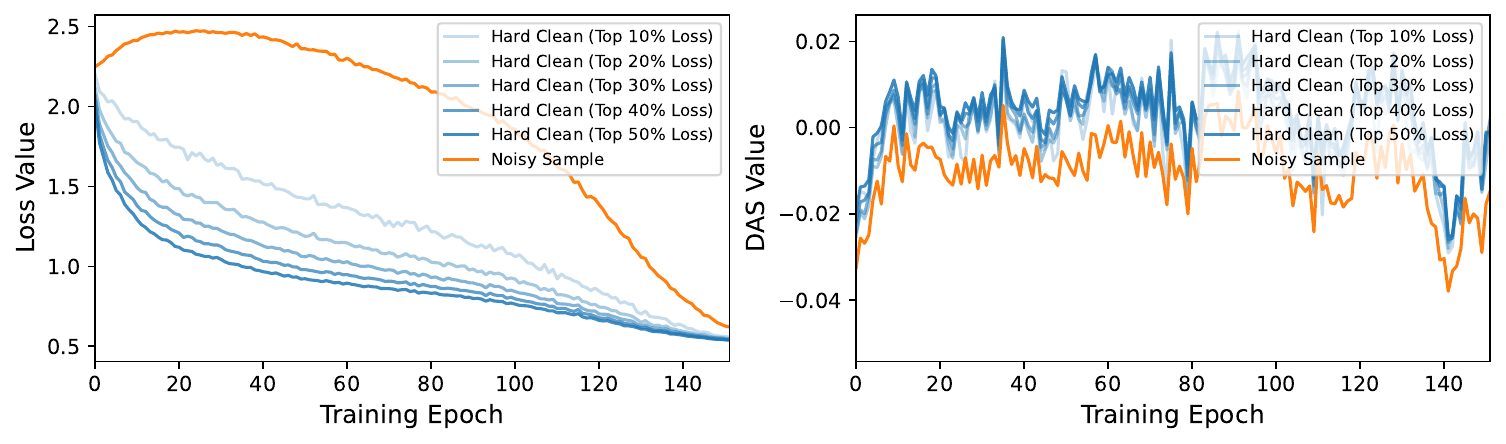}
    \vspace{-21pt}
    \caption{
    \textbf{Loss and DAS curves of hard \vs noisy samples.}
    Hard clean samples are identified as correctly labeled instances with the highest per-sample average loss across training, where~{\fontfamily{DejaVuSans-TLF}\selectfont\footnotesize Top x\% Loss} denotes the top x\% of clean samples by average loss. 
    %
    }
    \vspace{-8pt}
    \label{fig:vis_hard_vs_noisy}
\end{figure*}

\input{tab_supp/supp_results_pseudo_val_ablation}

\vspace{-6pt}
\paragraph{More Results with Pseudo Clean Data.}

To further demonstrate that AlignPrune does not strictly require any clean reference labels, we evaluate its performance using a pseudo-clean reference set estimated by one more recent label-free coreset method ELFS~\cite{zheng2025elfs}. 
As shown in~\Cref{tab:results_pseudo_val_more}, similarly consistent performance can be observed compared to using a clean validation reference set, further confirming AlignPrune's robustness in fully unsupervised noisy regimes.

\vspace{-6pt}
\paragraph{Discussion on Sensitivity to Reference Noise.} 

To address concerns regarding the purity of the reference set, we conduct a sensitivity analysis by injecting controlled synthetic noise directly into the reference. 
As shown in~\Cref{tab:reference_noise_sensitivity}, AlignPrune's performance on CIFAR-100N remains remarkably stable with reference noise ratios \textbf{up to 30\%}, dropping marginally for rates beyond 40\%, which further confirms the robustness of our trajectory alignment metric even under highly imperfect reference supervision.

\vspace{-6pt}
\paragraph{Remark.}

A broader study of distribution shift between the reference set and training distribution is orthogonal to our focus and is left to future work.

\subsubsection{Analysis on Effectiveness under Clean Labels}
\label{subsec:additional_analysis_effective_under_clean}

By definition, DAS of one sample measures the degree of alignment between its loss trajectory and the average trajectory of the reference set, reflecting whether the sample has the synchronized behavior as the clean reference samples.
Additionally, reference samples typically reflect the principal learning dynamics of the target task.
Thus, a sample with high DAS exhibits behavior that is more synchronized with principal learning dynamics, suggesting that it is more \textit{informative} and \textit{contributes meaningfully} to generalization.

Under clean-label scenario, this behavior implies that \algname{} naturally degrades into a surrogate of conventional dynamic pruning method.
In other words, when noise is absent, DAS serves as a proxy for identifying informative samples in a way that is consistent with established pruning methods such as InfoBatch or SeTa.
This dual behavior, robust under noisy-label and consistent under clean-label, supports the versatility of \algname{} across both noisy and noise-free settings.

\begin{table}[t]
    \centering
    \vspace{3pt}
    \caption{\textbf{Ablation on \algname{} with varying noise rates in reference.}
    We inject controlled noise into the reference to validate its noise sensitivity.
    }
    \vspace{-3pt}
    \footnotesize
    \setlength{\tabcolsep}{2.1pt}
    \adjustbox{width=\columnwidth}{
    \begin{tabular}{lccccc}
        \toprule
        \textbf{Ref.~Noise (\%)} $\rightarrow$ & 0\% & 10\% & 20\% & 30\% & 40\% \\
        \midrule
        Acc. ± std
        & 56.8\footnotesize{ ± 0.1} & 56.5\footnotesize{ ± 0.2} & 56.4\footnotesize{ ± 0.1} & 56.2\footnotesize{ ± 0.2} & 53.7\footnotesize{ ± 0.1} \\
        \bottomrule
    \end{tabular}
    }
    \vspace{-9pt}
    \label{tab:reference_noise_sensitivity}
\end{table}

\subsubsection{Analysis on Hard \textbf{\textit{vs.}} Noisy Samples}
\label{subsec:additional_analysis_hard_vs_noisy}

A critical challenge in robust learning is distinguishing \textit{hard-but-clean} samples from \textit{truly-noisy} ones, as both typically exhibit high loss values compared to \textit{easy-clean} samples. 
However, their learning dynamics differ fundamentally.
We plot the per-sample loss-value curves and DAS-value curves for these \textit{hard-clean} and \textit{noisy} samples in~\Cref{fig:vis_hard_vs_noisy}.
Hard clean samples are identified as correctly labeled instances with the highest per-sample average loss across training.

On the loss plots, loss values of hard clean samples are initially high, overlapping in magnitude with the noisy ones, which are nearly indistinguishable.
This confirms why loss-based methods fail in the noisy setting.
In contrast, the DAS curves clearly separate the two groups. 
Hard clean samples, despite their high loss, demonstrate a consistent and monotonic decrease that correlates positively with the reference trend, yielding persistently higher DAS.
Noisy samples exhibit a more erratic and inconsistent pattern, resulting in consistently low or negative correlations. 
These observations confirm that DAS is orthogonal to loss magnitude, enabling \algname{} to robustly retain \textit{hard-but-clean} samples while filtering out \textit{truly-noisy} ones.

\subsubsection{Analysis on Noise Ratio in Retained Subset}

\begin{figure}[t]
    \centering
    \includegraphics[width=\linewidth]{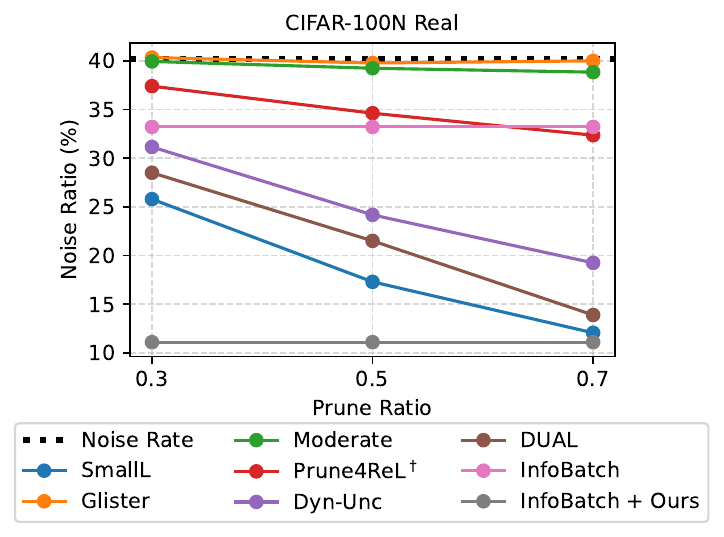}
    \vspace{-21pt}
    \caption{
    \textbf{Noise ratio in the retained subset on CIFAR-100N with Real noise-label.}
    Best viewed in color.
    }
    \vspace{-12pt}
    \label{fig:vis_noise_ratio_in_subset}
\end{figure}

To quantitatively verify that \algname{} effectively filters out noisy samples, we measure the fraction of noisy samples in the retained training subset across various pruning ratios.
\Cref{fig:vis_noise_ratio_in_subset} reports this fraction as noise ratio by each method on CIFAR-100N with 40.2\% Real noisy-label. 

As shown, the subset obtained by InfoBatch + Ours (\algname{}) yields a significantly lower noise ratio compared to both static baselines (SmallL, Prune4ReL) and dynamic baselines (InfoBatch). 
This improved subset purity confirms that the performance gains observed in our main results are attributable to the more precise identification and removal of noisy samples.

\subsubsection{Analysis on Preservation of Unbiasedness}

Existing dynamic pruning methods guarantee unbiased gradient estimation by applying expectation rescaling to the gradients of the retained samples, which is mathematically independent of how the subset is chosen. 
Our proposed \algname{} operates solely as a plug-and-play replacement for the ranking metric, thereby strictly preserving this unbiased gradient estimation guarantee.

\input{tab_supp/supp_results_statistic_analysis.tex}

\subsection{Additional Validation Beyond Image}

To further demonstrate the broad applicability of \algname{}, we evaluate it on the NEWS dataset from~\cite{kiryo2017positive} for text classification.
We follow the experimental settings in~\cite{yu2019does} to inject the symmetric and pairflip label noise with rates {0.2, 0.5} and 0.45, respectively.
As shown in~\cref{tab:supp_results_news}, \algname{} consistently outperforms baseline methods across various types of label noise, which confirms its effectiveness in different modalities beyond the image domain.

\input{tab_supp/supp_results_news}

\subsection{Additional Statistical Significance Analysis}

As mentioned in the experimental setup, each experiment is repeated three times, and we report the average accuracy across the runs.
Here, we provide a statistical significance analysis corresponding to the results in Tab.~\ref{tab:results_cifar100} under prune ratio of 30\%.
Results in~\cref{tab:supp_results_statistical_analysis} present the mean accuracy along with standard deviations, demonstrating the statistical reliability and stability of \algname{}.
These results confirm the consistency of performance gains observed with our approach.

%% file: tab_supp/supp_experiment_detail.tex
\begin{table*}[t]

\centering
\footnotesize
\setlength{\tabcolsep}{6pt}
\caption{\textbf{Summary of the experimental setups and hyper-parameters on CIFAR-100N, CIFAR-10N, WebVision and Clothing-1M datasets.}}
\begin{tabular}{lc|c|c|c|c}
\toprule
\multicolumn{2}{c|}{\multirow{1}{*}{\textbf{Hyper-parameters}}} & \multirow{1}{*}{\textbf{CIFAR-100N}} & \multirow{1}{*}{\textbf{CIFAR-10N}} & \multirow{1}{*}{\textbf{WebVision}} & \multirow{1}{*}{\textbf{Clothing-1M}} \\
\midrule
\multicolumn{1}{c|}{\multirow{7}{*}{\textbf{\begin{tabular}[c]{@{}c@{}}Training\\ Setup\end{tabular}}}} & architecture & ResNet-18 & ResNet-18 & InceptionResNetV2 & \begin{tabular}[c]{@{}c@{}}ResNet-50 \\ (Pre-trained)\end{tabular} \\
\multicolumn{1}{l|}{} & training epoch & 200 & 200 & 100 & 10 \\
\multicolumn{1}{l|}{} & batch size & 128 & 128 & 32 & 32 \\
\multicolumn{1}{l|}{} & optimizer & LARS & LARS & SGD & SGD \\
\multicolumn{1}{l|}{} & learning rate & 5.2 & 5.62 & 0.02 & 0.002 \\
\multicolumn{1}{l|}{} & lr scheduler & OneCycle & OneCycle & MultiStep-50th & MultiStep-5th \\
\multicolumn{1}{l|}{} & weight decay & $5 \times 10^{-4}$ & $5 \times 10^{-4}$ & $5 \times 10^{-4}$ & $5 \times 10^{-4}$ \\
\midrule
\multicolumn{1}{c|}{\multirow{2}{*}{\textbf{InfoBatch}}} & \textbf{$r$} & 0.5 & 0.5 & 0.75 & 0.75 \\
\multicolumn{1}{c|}{} & {$\delta$} & 0.875 & 0.875 & 0.875 & 0.875 \\
\midrule
\multicolumn{1}{c|}{\multirow{4}{*}{\textbf{SeTa}}}& \textbf{$r$} & 0.1 & 0.1 & \multirow{4}{*}{--} & \multirow{4}{*}{--} \\
\multicolumn{1}{c|}{} & \textbf{$\alpha$} & 0.9 & 0.9 &  & \\
\multicolumn{1}{c|}{} & \textbf{$k$} & 5 & 5 & & \\
\multicolumn{1}{c|}{} & {$\delta$} & 0.875 & 0.875 & & \\
\midrule
\multicolumn{1}{c|}{\multirow{2}{*}{\textbf{Ours}}} & \textbf{$N$} & 25 & 25 & 25 & 5 \\
\multicolumn{1}{c|}{} & {$\rho$} & Pearson & Pearson & Pearson & Pearson \\
\bottomrule
\end{tabular}
\label{table:supp_experiment_detail}
\vspace{-6pt}
\end{table*}

%% file: tab_supp/supp_results_cifar100.tex
\begin{table*}[htbp]
    \centering
    \footnotesize
    \setlength{\tabcolsep}{3pt}
    \vspace{-12pt}
    \caption{
    \textbf{Extended classification results on CIFAR-100N with ResNet-18.} 
    Performance gaps relative to the full-training setting are indicated as superscripts.
    Static pruning baselines are highlighted with gray.
    The mean $\Delta$ is computed across all noise types.
    $^\dagger$ indicates Prune4ReL with standard CE loss to ensure fair comparison on pure pruning methods.
    }
    \vspace{-6pt}
    \begin{tabular}{lcccccccc}
        \toprule
        Noisy Type $\rightarrow$ & \multirow{2}{*}{Clean} & \multirow{2}{*}{Real} & \multicolumn{3}{c}{Symmetric} & \multicolumn{2}{c}{Asymmetric} & Mean \\ 
        \cmidrule(lr){4-6}
        \cmidrule(lr){7-8}
        Pruning Method $\downarrow$ & & & 0.2 & 0.5 & 0.8 & 0.2 & 0.4 & $\Delta$ \\
        \midrule
        Full-training 
        & 78.2 & 56.1 & 71.4 & 58.6 & 39.8 & 72.4 & 63.3 & -- \\
        \midrule
        \multicolumn{9}{c}{\textbf{Prune Ratio $\sim$30\%}} \\
        \midrule
        \rowcolor{lightgray}
        Static Random 
        & 73.8\cgaphlr{-}{4.4} & 53.3\cgaphlr{-}{2.8} & 67.9\cgaphlr{-}{3.4} & 51.1\cgaphlr{-}{7.5} & 30.3\cgaphlr{-}{9.5} & 68.4\cgaphlr{-}{4.0} & 57.8\cgaphlr{-}{5.5} & -5.3 \\
        \rowcolor{lightgray}
        SmallL~\citep{jiang2018mentornet}
        & 71.3\cgaphlr{-}{6.9} & 58.0\cgaphl{+}{1.9} & 69.2\cgaphlr{-}{2.2} & 51.9\cgaphlr{-}{6.7} & 13.7\cgaphlr{-}{26.1} & 68.2\cgaphlr{-}{4.2} & 57.0\cgaphlr{-}{6.2} & -7.2 \\
        \rowcolor{lightgray}
        Margin~\citep{coleman2019selection}
        & 75.4\cgaphlr{-}{2.8} & 48.0\cgaphlr{-}{8.1} & 59.2\cgaphlr{-}{12.2} & 15.9\cgaphlr{-}{42.7} & 6.3\cgaphlr{-}{33.5} & 57.3\cgaphlr{-}{15.1} & 42.1\cgaphlr{-}{21.2} & -19.4 \\
        \rowcolor{lightgray}
        Forget~\citep{toneva2018empirical} 
        & 74.8\cgaphlr{-}{3.4} & 57.7\cgaphl{+}{1.6} & 70.2\cgaphlr{-}{1.1} & 51.3\cgaphlr{-}{7.3} & 12.5\cgaphlr{-}{27.4} & 68.4\cgaphlr{-}{4.0} & 59.2\cgaphlr{-}{4.1} & -6.5 \\
        \rowcolor{lightgray}
        GraNd~\citep{paul2021deep} 
        & 73.0\cgaphlr{-}{5.2} & 40.2\cgaphlr{-}{15.9} & 49.4\cgaphlr{-}{21.9} & 9.3\cgaphlr{-}{49.3} & 6.2\cgaphlr{-}{33.6} & 51.8\cgaphlr{-}{20.5} & 30.3\cgaphlr{-}{33.0} & -25.6 \\
        \rowcolor{lightgray}
        Glister~\cite{killamsetty2021glister}
        & 73.9\cgaphlr{-}{4.3} & 51.5\cgaphlr{-}{4.6} & 59.9\cgaphlr{-}{11.5} & 40.4\cgaphlr{-}{18.2} & 16.4\cgaphlr{-}{23.4} & 62.9\cgaphlr{-}{9.5} & 49.7\cgaphlr{-}{13.6} & -12.2 \\
        \rowcolor{lightgray}
        Moderate~\citep{xia2022moderate} 
        & 73.2\cgaphlr{-}{5.0} & 52.1\cgaphlr{-}{4.0} & 60.2\cgaphlr{-}{11.2} & 30.2\cgaphlr{-}{28.4} & 9.9\cgaphlr{-}{29.9} & 60.0\cgaphlr{-}{12.4} & 45.7\cgaphlr{-}{17.6} & -15.5 \\ 
        \rowcolor{lightgray}
        SSP~\citep{sorscher2022beyond} 
        & 74.9\cgaphlr{-}{3.3} & 53.0\cgaphlr{-}{3.1} & 60.3\cgaphlr{-}{11.1} & 35.8\cgaphlr{-}{22.9} & 7.9\cgaphlr{-}{31.9} & 62.9\cgaphlr{-}{9.5} & 50.4\cgaphlr{-}{12.8} & -13.5 \\
        \rowcolor{lightgray}
        Prune4ReL$^\dagger$~\citep{park2023robust} 
        & 73.2\cgaphlr{-}{5.0} & 52.3\cgaphlr{-}{3.8} & 61.2\cgaphlr{-}{10.1} & 37.0\cgaphlr{-}{21.6} & 9.0\cgaphlr{-}{30.8} & 61.1\cgaphlr{-}{11.3} & 48.9\cgaphlr{-}{14.4} & -13.8 \\
        \rowcolor{lightgray}
        Dyn-Unc~\citep{he2024large}
        & 77.2\cgaphlr{-}{1.0} & 58.5\cgaphl{+}{2.3} & 70.9\cgaphlr{-}{0.5} & 55.4\cgaphlr{-}{3.2} & 24.3\cgaphlr{-}{15.5} & 71.8\cgaphlr{-}{0.6} & 62.6\cgaphlr{-}{0.7} & -2.7 \\
        \rowcolor{lightgray}
        DUAL~\citep{cho2025lightweight}
        & 78.1\cgaphlr{-}{0.1} & 58.8\cgaphl{+}{2.7} & 71.3\cgaphlr{-}{0.1} & 55.4\cgaphlr{-}{3.2} & 22.7\cgaphlr{-}{17.1} & 72.2\cgaphlr{-}{0.2} & 60.8\cgaphlr{-}{2.4} & -2.9 \\
        Dynamic Random~\cite{okanovic2024repeated}  
        & 77.3\cgaphlr{-}{0.9} & 54.7\cgaphlr{-}{1.4} & 69.9\cgaphlr{-}{1.4} & 58.8\cgaphl{+}{0.2} & 40.1\cgaphl{+}{0.3} & 71.5\cgaphlr{-}{0.9} & 63.2\cgaphlr{-}{0.1} & -0.6 \\
        DivBS~\citep{hong2024diversified}
        & 77.1\cgaphlr{-}{1.1} & 54.6\cgaphlr{-}{1.5} & 63.5\cgaphlr{-}{7.9} & 51.0\cgaphlr{-}{7.6} & 20.8\cgaphlr{-}{19.0} & 65.9\cgaphlr{-}{6.5} & 56.2\cgaphlr{-}{7.1} & -7.2 \\
        IES~\citep{yuan2025instance}
        & 76.1\cgaphlr{-}{2.1} & 53.9\cgaphlr{-}{2.2} & 64.2\cgaphlr{-}{7.2} & 51.0\cgaphlr{-}{7.6} & 37.0\cgaphlr{-}{2.8} & 65.3\cgaphlr{-}{7.0} & 53.8\cgaphlr{-}{9.5} & -5.5 \\
        InfoBatch~\citep{qin2024infobatch}
        & 79.0\cgaphl{+}{0.8} & 56.1\cgap{+}{0.0} & 71.4\cgap{+}{0.0} & 59.7\cgaphl{+}{1.1} & 41.8\cgaphl{+}{2.0} & 71.9\cgaphlr{-}{0.5} & 64.2\cgaphl{+}{0.9} & +0.6 \\
        \textbf{InfoBatch + Ours} 
        & \cellcolor{cyan!20}79.3\cgaphl{+}{1.1} & \cellcolor{cyan!20}59.4\cgaphl{+}{3.3} & \cellcolor{cyan!20}71.8\cgaphl{+}{0.4} & \cellcolor{cyan!20}66.0\cgaphl{+}{7.4} & \cellcolor{cyan!20}41.8\cgaphl{+}{2.0} & \cellcolor{cyan!20}72.6\cgaphl{+}{0.2} & \cellcolor{cyan!20}68.0\cgaphl{+}{4.7} & \cellcolor{cyan!20}\textbf{+2.7} \\
        SeTa~\citep{zhou2025scale} 
        & 79.0\cgaphl{+}{0.8} & 55.6\cgaphlr{-}{0.5} & 70.2\cgaphlr{-}{1.2} & 59.0\cgaphl{+}{0.4} & 41.6\cgaphl{+}{1.8} & 71.4\cgaphlr{-}{0.9} & 63.2\cgaphlr{-}{0.1} & +0.0 \\
        \textbf{SeTa + Ours}  
        & \cellcolor{cyan!20}79.3\cgaphl{+}{1.1} & \cellcolor{cyan!20}56.3\cgaphl{+}{0.2} & \cellcolor{cyan!20}70.8\cgaphlr{-}{0.5} & \cellcolor{cyan!20}60.5\cgaphl{+}{1.9} & \cellcolor{cyan!20}41.6\cgaphl{+}{1.8} & \cellcolor{cyan!20}71.9\cgaphlr{-}{0.5} & \cellcolor{cyan!20}64.3\cgaphl{+}{1.0} & \cellcolor{cyan!20}\textbf{+0.7} \\
        \midrule
        \multicolumn{9}{c}{\textbf{Prune Ratio $\sim$50\%}} \\
        \midrule
        \rowcolor{lightgray}
        Static Random 
        & 72.1\cgaphlr{-}{6.1} & 51.8\cgaphlr{-}{4.3} & 64.0\cgaphlr{-}{7.4} & 47.2\cgaphlr{-}{11.5} & 22.9\cgaphlr{-}{16.9} & 65.3\cgaphlr{-}{7.0} & 54.6\cgaphlr{-}{8.7} & -8.8 \\
        \rowcolor{lightgray}
        SmallL~\citep{jiang2018mentornet}
        & 65.8\cgaphlr{-}{12.4} & 56.1\cgaphlr{-}{0.1} & 64.9\cgaphlr{-}{6.4} & 57.9\cgaphlr{-}{0.8} & 18.0\cgaphlr{-}{21.8} & 64.2\cgaphlr{-}{8.1} & 58.8\cgaphlr{-}{4.4} & -7.7 \\
        \rowcolor{lightgray}
        Margin~\citep{coleman2019selection}
        & 71.2\cgaphlr{-}{7.0} & 36.3\cgaphlr{-}{19.8} & 44.7\cgaphlr{-}{26.7} & 11.0\cgaphlr{-}{47.6} & 5.9\cgaphlr{-}{33.9} & 43.7\cgaphlr{-}{28.7} & 26.8\cgaphlr{-}{36.5} & -28.6 \\
        \rowcolor{lightgray}
        Forget~\citep{toneva2018empirical} 
        & 69.8\cgaphlr{-}{8.4} & 56.4\cgaphl{+}{0.3} & 68.4\cgaphlr{-}{2.9} & 60.4\cgaphl{+}{1.8} & 16.5\cgaphlr{-}{23.3} & 66.1\cgaphlr{-}{6.3} & 60.7\cgaphlr{-}{2.5} & -5.9 \\
        \rowcolor{lightgray}
        GraNd~\citep{paul2021deep} 
        & 64.0\cgaphlr{-}{14.2} & 25.8\cgaphlr{-}{30.3} & 28.6\cgaphlr{-}{42.8} & 4.1\cgaphlr{-}{54.5} & 4.3\cgaphlr{-}{35.5} & 36.6\cgaphlr{-}{35.8} & 18.2\cgaphlr{-}{45.1} & -36.9 \\
        \rowcolor{lightgray}
        Glister~\cite{killamsetty2021glister}
        & 69.5\cgaphlr{-}{8.8} & 49.2\cgaphlr{-}{6.9} & 55.0\cgaphlr{-}{16.3} & 34.3\cgaphlr{-}{24.4} & 12.9\cgaphlr{-}{26.9} & 58.2\cgaphlr{-}{14.2} & 45.0\cgaphlr{-}{18.3} & -16.5 \\
        \rowcolor{lightgray}
        Moderate~\citep{xia2022moderate} 
        & 68.7\cgaphlr{-}{9.5} & 49.0\cgaphlr{-}{7.2} & 52.7\cgaphlr{-}{18.6} & 22.3\cgaphlr{-}{36.3} & 6.4\cgaphlr{-}{33.4} & 55.1\cgaphlr{-}{17.3} & 41.0\cgaphlr{-}{22.3} & -20.6 \\ 
        \rowcolor{lightgray}
        SSP~\citep{sorscher2022beyond} 
        & 70.6\cgaphlr{-}{7.6} & 49.2\cgaphlr{-}{6.9} & 52.9\cgaphlr{-}{18.5} & 31.1\cgaphlr{-}{27.5} & 7.5\cgaphlr{-}{32.3} & 58.6\cgaphlr{-}{13.8} & 47.5\cgaphlr{-}{15.8} & -17.5 \\
        \rowcolor{lightgray}
        Prune4ReL$^\dagger$~\citep{park2023robust} 
        & 69.1\cgaphlr{-}{9.1} & 51.7\cgaphlr{-}{4.4} & 58.2\cgaphlr{-}{13.2} & 32.7\cgaphlr{-}{25.9} & 7.4\cgaphlr{-}{32.4} & 58.7\cgaphlr{-}{13.7} & 45.3\cgaphlr{-}{17.9} & -16.7 \\
        \rowcolor{lightgray}
        Dyn-Unc~\citep{he2024large}
        & 74.1\cgaphlr{-}{4.2} & 59.7\cgaphl{+}{3.5} & 71.4\cgap{+}{0.0} & 61.7\cgaphl{+}{3.1} & 22.8\cgaphlr{-}{17.0} & 71.1\cgaphlr{-}{1.2} & 66.5\cgaphl{+}{3.2} & -1.8 \\
        \rowcolor{lightgray}
        DUAL~\citep{cho2025lightweight}
        & 74.6\cgaphlr{-}{3.6} & 59.9\cgaphl{+}{3.8} & 71.3\cgaphlr{-}{0.1} & 61.9\cgaphl{+}{3.3} & 20.6\cgaphlr{-}{19.2} & 71.1\cgaphlr{-}{1.3} & 62.9\cgaphlr{-}{0.4} & -2.5 \\
        Dynamic Random~\cite{okanovic2024repeated}  
        & 75.3\cgaphlr{-}{2.9} & 54.1\cgaphlr{-}{2.0} & 70.4\cgaphlr{-}{1.0} & 59.5\cgaphl{+}{0.9} & 40.7\cgaphl{+}{0.9} & 70.1\cgaphlr{-}{2.3} & 64.8\cgaphl{+}{1.6} & -0.7 \\
        DivBS~\citep{hong2024diversified}
        & 77.1\cgaphlr{-}{1.1} & 53.8\cgaphlr{-}{2.3} & 63.8\cgaphlr{-}{7.5} & 46.6\cgaphlr{-}{12.0} & 18.0\cgaphlr{-}{21.9} & 64.5\cgaphlr{-}{7.8} & 56.7\cgaphlr{-}{6.6} & -8.5 \\
        IES~\citep{yuan2025instance}
        & 75.5\cgaphlr{-}{2.7} & 53.2\cgaphlr{-}{2.9} & 62.7\cgaphlr{-}{8.7} & 50.2\cgaphlr{-}{8.4} & 34.4\cgaphlr{-}{5.4} & 63.3\cgaphlr{-}{9.1} & 54.8\cgaphlr{-}{8.5} & -6.5 \\
        InfoBatch~\citep{qin2024infobatch}
        & 77.7\cgaphlr{-}{0.5} & 56.0\cgaphlr{-}{0.1} & 71.3\cgaphlr{-}{0.1} & 60.5\cgaphl{+}{1.9} & 42.2\cgaphl{+}{2.4} & 71.8\cgaphlr{-}{0.6} & 65.2\cgaphl{+}{2.0} & +0.7 \\
        \textbf{InfoBatch + Ours}  
        & \cellcolor{cyan!20}78.5\cgaphl{+}{0.3} & \cellcolor{cyan!20}60.7\cgaphl{+}{4.6} & \cellcolor{cyan!20}71.6\cgaphl{+}{0.3} & \cellcolor{cyan!20}62.0\cgaphl{+}{3.4} & \cellcolor{cyan!20}42.6\cgaphl{+}{2.8} & \cellcolor{cyan!20}72.6\cgaphl{+}{0.3} & \cellcolor{cyan!20}68.6\cgaphl{+}{5.3} & \cellcolor{cyan!20}\textbf{+2.4} \\
        SeTa~\citep{zhou2025scale} 
        & 77.5\cgaphlr{-}{0.7} & 55.7\cgaphlr{-}{0.4} & 70.7\cgaphlr{-}{0.7} & 60.0\cgaphl{+}{1.4} & 40.5\cgaphl{+}{0.7} & 71.9\cgaphlr{-}{0.5} & 64.5\cgaphl{+}{1.2} & +0.2 \\
        \textbf{SeTa + Ours} 
        & \cellcolor{cyan!20}78.4\cgaphl{+}{0.2} & \cellcolor{cyan!20}56.3\cgaphl{+}{0.1} & \cellcolor{cyan!20}71.2\cgaphlr{-}{0.2} & \cellcolor{cyan!20}61.0\cgaphl{+}{2.4} & \cellcolor{cyan!20}41.9\cgaphl{+}{2.1} & \cellcolor{cyan!20}72.2\cgaphlr{-}{0.2} & \cellcolor{cyan!20}66.0\cgaphl{+}{2.7} & \cellcolor{cyan!20}\textbf{+1.0} \\
        \midrule
        \multicolumn{9}{c}{\textbf{Prune Ratio $\sim$70\%}} \\
        \midrule
        \rowcolor{lightgray}
        Static Random 
        & 69.7\cgaphlr{-}{8.5} & 45.9\cgaphlr{-}{10.2} & 56.7\cgaphlr{-}{14.6} & 39.0\cgaphlr{-}{19.6} & 15.8\cgaphlr{-}{24.0} & 58.4\cgaphlr{-}{13.9} & 49.5\cgaphlr{-}{13.8} & -14.9 \\
        \rowcolor{lightgray}
        SmallL~\citep{jiang2018mentornet}
        & 55.7\cgaphlr{-}{22.5} & 49.8\cgaphlr{-}{6.3} & 57.0\cgaphlr{-}{14.4} & 54.6\cgaphlr{-}{4.1} & 22.0\cgaphlr{-}{17.8} & 56.7\cgaphlr{-}{15.7} & 53.8\cgaphlr{-}{9.5} & -12.9 \\
        \rowcolor{lightgray}
        Margin~\citep{coleman2019selection}
        & 56.7\cgaphlr{-}{21.5} & 10.6\cgaphlr{-}{45.5} & 12.3\cgaphlr{-}{59.1} & 6.5\cgaphlr{-}{52.2} & 4.5\cgaphlr{-}{35.3} & 17.1\cgaphlr{-}{55.3} & 5.2\cgaphlr{-}{58.1} & -46.7 \\
        \rowcolor{lightgray}
        Forget~\citep{toneva2018empirical} 
        & 58.6\cgaphlr{-}{19.6} & 50.0\cgaphlr{-}{6.1} & 61.8\cgaphlr{-}{9.6} & 59.3\cgaphl{+}{0.7} & 22.8\cgaphlr{-}{17.0} & 58.4\cgaphlr{-}{14.0} & 56.1\cgaphlr{-}{7.2} & -10.4 \\
        \rowcolor{lightgray}
        GraNd~\citep{paul2021deep} 
        & 46.2\cgaphlr{-}{32.0} & 14.7\cgaphlr{-}{41.4} & 10.1\cgaphlr{-}{61.3} & 2.9\cgaphlr{-}{55.8} & 3.4\cgaphlr{-}{36.4} & 20.1\cgaphlr{-}{52.3} & 9.1\cgaphlr{-}{54.2} & -47.6 \\
        \rowcolor{lightgray}
        Glister~\cite{killamsetty2021glister}
        & 61.1\cgaphlr{-}{17.1} & 43.6\cgaphlr{-}{12.5} & 46.1\cgaphlr{-}{25.3} & 22.4\cgaphlr{-}{36.2} & 9.1\cgaphlr{-}{30.7} & 49.9\cgaphlr{-}{22.5} & 38.5\cgaphlr{-}{24.8} & -24.1 \\
        \rowcolor{lightgray}
        Moderate~\citep{xia2022moderate} 
        & 60.6\cgaphlr{-}{17.6} & 43.3\cgaphlr{-}{12.8} & 46.2\cgaphlr{-}{25.2} & 17.7\cgaphlr{-}{41.0} & 5.1\cgaphlr{-}{34.7} & 46.6\cgaphlr{-}{25.8} & 36.5\cgaphlr{-}{26.7} & -26.3 \\ 
        \rowcolor{lightgray}
        SSP~\citep{sorscher2022beyond} 
        & 60.3\cgaphlr{-}{17.9} & 42.2\cgaphlr{-}{14.0} & 41.3\cgaphlr{-}{30.1} & 21.6\cgaphlr{-}{37.0} & 6.1\cgaphlr{-}{33.8} & 49.2\cgaphlr{-}{23.2} & 37.7\cgaphlr{-}{25.6} & -25.9 \\
        \rowcolor{lightgray}
        Prune4ReL$^\dagger$~\citep{park2023robust} 
        & 61.4\cgaphlr{-}{16.8} & 46.2\cgaphlr{-}{10.0} & 51.1\cgaphlr{-}{20.3} & 28.0\cgaphlr{-}{30.7} & 6.1\cgaphlr{-}{33.7} & 52.4\cgaphlr{-}{20.0} & 42.8\cgaphlr{-}{20.5} & -21.7 \\
        \rowcolor{lightgray}
        Dyn-Unc~\citep{he2024large}
        & 63.9\cgaphlr{-}{14.3} & 57.5\cgaphl{+}{1.4} & 67.1\cgaphlr{-}{4.3} & 63.4\cgaphl{+}{4.7} & 24.6\cgaphlr{-}{15.2} & 65.8\cgaphlr{-}{6.6} & 63.0\cgaphlr{-}{0.2} & -4.9 \\
        \rowcolor{lightgray}
        DUAL~\citep{cho2025lightweight}
        & 68.9\cgaphlr{-}{9.3} & 57.9\cgaphl{+}{1.7} & 66.4\cgaphlr{-}{5.0} & 62.5\cgaphl{+}{3.9} & 19.9\cgaphlr{-}{19.9} & 67.3\cgaphlr{-}{5.1} & 61.5\cgaphlr{-}{1.8} & -5.1 \\
        Dynamic Random~\cite{okanovic2024repeated}  
        & 75.2\cgaphlr{-}{3.0} & 53.3\cgaphlr{-}{2.9} & 70.2\cgaphlr{-}{1.2} & 62.7\cgaphl{+}{4.1} & 41.0\cgaphl{+}{1.2} & 71.9\cgaphlr{-}{0.5} & 67.3\cgaphl{+}{4.0} & +0.3 \\
        DivBS~\citep{hong2024diversified}
        & 76.9\cgaphlr{-}{1.3} & 53.6\cgaphlr{-}{2.6} & 61.5\cgaphlr{-}{9.8} & 45.8\cgaphlr{-}{12.5} & 19.4\cgaphlr{-}{20.4} & 63.5\cgaphlr{-}{8.8} & 58.5\cgaphlr{-}{4.8} & -8.6 \\
        IES~\citep{yuan2025instance}
        & 74.8\cgaphlr{-}{3.4} & 52.6\cgaphlr{-}{3.6} & 61.0\cgaphlr{-}{10.4} & 50.3\cgaphlr{-}{8.3} & 28.6\cgaphlr{-}{11.2} & 62.6\cgaphlr{-}{9.8} & 51.9\cgaphlr{-}{11.4} & -8.3 \\
        InfoBatch~\citep{qin2024infobatch} 
        & 77.1\cgaphlr{-}{1.1} & 55.0\cgaphlr{-}{1.2} & 71.4\cgaphl{+}{0.1} & 63.6\cgaphl{+}{5.0} & 42.4\cgaphl{+}{2.6} & 72.2\cgaphlr{-}{0.2} & 67.8\cgaphl{+}{4.5} & +1.4 \\
        \textbf{InfoBatch + Ours} 
        & \cellcolor{cyan!20}77.5\cgaphlr{-}{0.7} & \cellcolor{cyan!20}58.3\cgaphl{+}{2.2} & \cellcolor{cyan!20}72.2\cgaphl{+}{0.9} & \cellcolor{cyan!20}64.7\cgaphl{+}{6.1} & \cellcolor{cyan!20}42.8\cgaphl{+}{3.0} & \cellcolor{cyan!20}72.5\cgaphl{+}{0.1} & \cellcolor{cyan!20}69.1\cgaphl{+}{5.9} & \cellcolor{cyan!20}\textbf{+2.5} \\
        SeTa~\citep{zhou2025scale}
        & 77.2\cgaphlr{-}{1.1} & 55.2\cgaphlr{-}{1.0} & 71.6\cgaphl{+}{0.2} & 62.4\cgaphl{+}{3.8} & 42.4\cgaphl{+}{2.6} & 72.0\cgaphlr{-}{0.4} & 67.4\cgaphl{+}{4.1} & +1.2 \\
        \textbf{SeTa + Ours} 
        & \cellcolor{cyan!20}77.8\cgaphlr{-}{0.4} & \cellcolor{cyan!20}55.5\cgaphlr{-}{0.6} & \cellcolor{cyan!20}72.3\cgaphl{+}{0.9} & \cellcolor{cyan!20}63.3\cgaphl{+}{4.7} & \cellcolor{cyan!20}42.7\cgaphl{+}{2.9} & \cellcolor{cyan!20}72.7\cgaphl{+}{0.3} & \cellcolor{cyan!20}67.6\cgaphl{+}{4.3} & \cellcolor{cyan!20}\textbf{+1.7} \\
        \bottomrule
    \end{tabular}
    \vspace{-6pt}
    \label{tab:supp_results_cifar100_full}
\end{table*}

%% file: tab_supp/supp_results_cifar10.tex
\begin{table*}[htbp]
    \centering
    \footnotesize
    \setlength{\tabcolsep}{3pt}
    \vspace{-12pt}
    \caption{\textbf{Extended classification results on CIFAR-10N with ResNet-18.} 
    Performance gaps relative to the full-training setting are indicated as superscripts.
    Static pruning methods are highlighted with gray.
    The mean $\Delta$ is computed across all noise types.
    $^\dagger$ indicates Prune4ReL with standard CE loss to ensure fair comparison on pure pruning methods.
    }
    \vspace{-6pt}
    \begin{tabular}{lccccccccc}
        \toprule
        Noisy Type $\rightarrow$ & \multirow{2}{*}{Clean} & \multicolumn{2}{c}{Real} & \multicolumn{3}{c}{Symmetric} & \multicolumn{2}{c}{Asymmetric} & Mean \\ 
        \cmidrule(lr){3-4}
        \cmidrule(lr){5-7}
        \cmidrule(lr){8-9}
        Pruning Method $\downarrow$ & & Real-A & Real-W & 0.2 & 0.5 & 0.8 & 0.2 & 0.4 & $\Delta$ \\
        \midrule
        Full-training 
        & 95.6 & 90.7 & 78.3 & 91.3 & 85.4 & 65.6 & 90.6 & 85.8 & -- \\
        \midrule
        \multicolumn{10}{c}{\textbf{Prune Ratio $\sim$30\%}} \\
        \midrule
        \rowcolor{lightgray}
        Static Random 
        & 94.6\cgaphlr{-}{1.0} & 89.9\cgaphlr{-}{0.7} & 75.0\cgaphlr{-}{3.3} & 89.0\cgaphlr{-}{2.3} & 81.2\cgaphlr{-}{4.2} & 57.6\cgaphlr{-}{8.0} & 87.9\cgaphlr{-}{2.8} & 82.5\cgaphlr{-}{3.3} & -3.2 \\
        \rowcolor{lightgray}
        SmallL~\citep{jiang2018mentornet}
        & 88.9\cgaphlr{-}{6.7} & 87.2\cgaphlr{-}{3.5} & 78.2\cgaphlr{-}{0.1} & 89.3\cgaphlr{-}{2.0} & 72.1\cgaphlr{-}{13.3} & 24.0\cgaphlr{-}{41.6} & 87.4\cgaphlr{-}{3.3} & 68.9\cgaphlr{-}{16.9} & -10.9 \\
        \rowcolor{lightgray}
        Margin~\citep{coleman2019selection}
        & 94.5\cgaphlr{-}{1.1} & 88.9\cgaphlr{-}{1.8} & 64.6\cgaphlr{-}{13.7} & 85.4\cgaphlr{-}{5.9} & 15.0\cgaphlr{-}{70.5} & 12.0\cgaphlr{-}{53.6} & 79.8\cgaphlr{-}{10.8} & 47.6\cgaphlr{-}{38.2} & -24.4 \\
        \rowcolor{lightgray}
        Forget~\citep{toneva2018empirical}
        & 93.6\cgaphlr{-}{2.0} & 89.5\cgaphlr{-}{1.1} & 69.6\cgaphlr{-}{8.8} & 84.5\cgaphlr{-}{6.8} & 59.0\cgaphlr{-}{26.4} & 22.5\cgaphlr{-}{43.1} & 80.5\cgaphlr{-}{10.2} & 58.3\cgaphlr{-}{27.5} & -15.7 \\
        \rowcolor{lightgray}
        GraNd~\citep{paul2021deep}
        & 93.9\cgaphlr{-}{1.7} & 87.8\cgaphlr{-}{2.9} & 40.5\cgaphlr{-}{37.8} & 65.9\cgaphlr{-}{25.4} & 17.5\cgaphlr{-}{68.0} & 11.0\cgaphlr{-}{54.6} & 60.5\cgaphlr{-}{30.1} & 38.7\cgaphlr{-}{47.1} & -33.4 \\
        \rowcolor{lightgray}
        Glister~\cite{killamsetty2021glister}
        & 93.1\cgaphlr{-}{2.5} & 88.7\cgaphlr{-}{2.0} & 75.6\cgaphlr{-}{2.8} & 83.8\cgaphlr{-}{7.5} & 73.1\cgaphlr{-}{12.4} & 37.1\cgaphlr{-}{28.6} & 84.1\cgaphlr{-}{6.5} & 74.3\cgaphlr{-}{11.5} & -9.2 \\
        \rowcolor{lightgray}
        Moderate~\citep{xia2022moderate}
        & 92.1\cgaphlr{-}{3.5} & 89.9\cgaphlr{-}{0.7} & 66.8\cgaphlr{-}{11.5} & 87.8\cgaphlr{-}{3.5} & 52.8\cgaphlr{-}{32.6} & 17.8\cgaphlr{-}{47.9} & 80.8\cgaphlr{-}{9.9} & 58.3\cgaphlr{-}{27.5} & -17.1 \\ 
        \rowcolor{lightgray}
        SSP~\citep{sorscher2022beyond}
        & 93.4\cgaphlr{-}{2.2} & 88.6\cgaphlr{-}{2.1} & 66.0\cgaphlr{-}{12.3} & 83.4\cgaphlr{-}{7.9} & 53.1\cgaphlr{-}{32.4} & 16.6\cgaphlr{-}{49.0} & 80.3\cgaphlr{-}{10.4} & 59.5\cgaphlr{-}{26.3} & -17.8 \\
        \rowcolor{lightgray}
        Prune4ReL$^\dagger$~\citep{park2023robust}
        & 93.0\cgaphlr{-}{2.6} & 88.3\cgaphlr{-}{2.4} & 66.1\cgaphlr{-}{12.3} & 81.8\cgaphlr{-}{9.5} & 52.2\cgaphlr{-}{33.2} & 15.8\cgaphlr{-}{49.8} & 81.1\cgaphlr{-}{9.6} & 59.0\cgaphlr{-}{26.8} & -18.3 \\
        \rowcolor{lightgray}
        Dyn-Unc~\citep{he2024large}
        & 95.3\cgaphlr{-}{0.3} & 89.6\cgaphlr{-}{1.0} & 78.7\cgaphl{+}{0.4} & 89.6\cgaphlr{-}{1.7} & 78.6\cgaphlr{-}{6.9} & 44.8\cgaphlr{-}{20.8} & 82.6\cgaphlr{-}{8.0} & 76.1\cgaphlr{-}{9.7} & -6.0 \\
        \rowcolor{lightgray}
        DUAL~\citep{cho2025lightweight}
        & 95.5\cgaphlr{-}{0.1} & 90.1\cgaphlr{-}{0.5} & 80.4\cgaphl{+}{2.0} & 91.7\cgaphl{+}{0.4} & 79.2\cgaphlr{-}{6.2} & 40.2\cgaphlr{-}{25.5} & 87.1\cgaphlr{-}{3.6} & 67.8\cgaphlr{-}{18.0} & -6.4 \\
        Dynamic Random~\cite{okanovic2024repeated}  
        & 94.8\cgaphlr{-}{0.8} & 90.3\cgaphlr{-}{0.3} & 77.3\cgaphlr{-}{1.0} & 90.8\cgaphlr{-}{0.5} & 84.2\cgaphlr{-}{1.2} & 64.8\cgaphlr{-}{0.8} & 91.5\cgaphl{+}{0.9} & 84.5\cgaphlr{-}{1.3} & -0.6 \\
        DivBS~\citep{hong2024diversified}
        & 95.3\cgaphlr{-}{0.3} & 90.3\cgaphlr{-}{0.3} & 78.3\cgap{+}{0.0} & 85.3\cgaphlr{-}{6.0} & 79.2\cgaphlr{-}{6.3} & 50.3\cgaphlr{-}{15.3} & 88.0\cgaphlr{-}{2.6} & 82.8\cgaphlr{-}{3.0} & -4.2 \\
        IES~\citep{yuan2025instance}
        & 95.0\cgaphlr{-}{0.6} & 89.8\cgaphlr{-}{0.8} & 79.8\cgaphl{+}{1.5} & 87.5\cgaphlr{-}{3.8} & 80.0\cgaphlr{-}{5.4} & 46.0\cgaphlr{-}{19.7} & 88.9\cgaphlr{-}{1.8} & 82.3\cgaphlr{-}{3.5} & -4.3 \\
        InfoBatch~\citep{qin2024infobatch} 
        & 95.3\cgaphlr{-}{0.3} & 91.0\cgaphl{+}{0.3} & 78.1\cgaphlr{-}{0.3} & 92.0\cgaphl{+}{0.7} & 85.6\cgaphl{+}{0.1} & 67.5\cgaphl{+}{1.9} & 91.3\cgaphl{+}{0.7} & 87.0\cgaphl{+}{1.3} & +0.5 \\
        \textbf{InfoBatch + Ours} 
        & \cellcolor{cyan!20}95.3\cgaphlr{-}{0.3} & \cellcolor{cyan!20}91.4\cgaphl{+}{0.7} & \cellcolor{cyan!20}82.0\cgaphl{+}{3.7} & \cellcolor{cyan!20}92.7\cgaphl{+}{1.4} & \cellcolor{cyan!20}87.9\cgaphl{+}{2.5} & \cellcolor{cyan!20}67.8\cgaphl{+}{2.2} & \cellcolor{cyan!20}92.9\cgaphl{+}{2.3} & \cellcolor{cyan!20}90.0\cgaphl{+}{4.2} & \cellcolor{cyan!20}\textbf{+2.1} \\
        SeTa~\citep{zhou2025scale}
        & 95.2\cgaphlr{-}{0.4} & 90.6\cgaphlr{-}{0.1} & 76.3\cgaphlr{-}{2.0} & 91.8\cgaphl{+}{0.5} & 84.7\cgaphlr{-}{0.7} & 64.2\cgaphlr{-}{1.5} & 90.8\cgaphl{+}{0.1} & 88.4\cgaphl{+}{2.6} & -0.2 \\
        \textbf{SeTa + Ours}  
        & \cellcolor{cyan!20}95.3\cgaphlr{-}{0.3} & \cellcolor{cyan!20}90.7\cgap{+}{0.0} & \cellcolor{cyan!20}79.1\cgaphl{+}{0.8} & \cellcolor{cyan!20}92.5\cgaphl{+}{1.2} & \cellcolor{cyan!20}85.0\cgaphlr{-}{0.5} & \cellcolor{cyan!20}64.6\cgaphlr{-}{1.0} & \cellcolor{cyan!20}91.8\cgaphl{+}{1.2} & \cellcolor{cyan!20}89.3\cgaphl{+}{3.5} & \cellcolor{cyan!20}\textbf{+0.6} \\
        \midrule
        \multicolumn{10}{c}{\textbf{Prune Ratio $\sim$50\%}} \\
        \midrule
        \rowcolor{lightgray}
        Static Random 
        & 93.3\cgaphlr{-}{2.3} & 88.6\cgaphlr{-}{2.1} & 72.4\cgaphlr{-}{5.9} & 87.4\cgaphlr{-}{3.9} & 77.5\cgaphlr{-}{8.0} & 51.4\cgaphlr{-}{14.2} & 85.8\cgaphlr{-}{4.9} & 77.9\cgaphlr{-}{7.9} & -6.1 \\
        \rowcolor{lightgray}
        SmallL~\citep{jiang2018mentornet}
        & 84.1\cgaphlr{-}{11.5} & 84.1\cgaphlr{-}{6.6} & 78.0\cgaphlr{-}{0.3} & 84.7\cgaphlr{-}{6.6} & 80.8\cgaphlr{-}{4.6} & 30.9\cgaphlr{-}{34.7} & 84.7\cgaphlr{-}{6.0} & 60.9\cgaphlr{-}{24.9} & -11.9 \\
        \rowcolor{lightgray}
        Margin~\citep{coleman2019selection}
        & 93.3\cgaphlr{-}{2.3} & 85.2\cgaphlr{-}{5.5} & 44.6\cgaphlr{-}{33.8} & 70.7\cgaphlr{-}{20.7} & 13.5\cgaphlr{-}{72.0} & 10.7\cgaphlr{-}{54.9} & 57.5\cgaphlr{-}{33.1} & 38.6\cgaphlr{-}{47.2} & -33.7 \\
        \rowcolor{lightgray}
        Forget~\citep{toneva2018empirical}
        & 92.4\cgaphlr{-}{3.2} & 88.4\cgaphlr{-}{2.3} & 72.2\cgaphlr{-}{6.1} & 85.7\cgaphlr{-}{5.6} & 68.3\cgaphlr{-}{17.2} & 29.8\cgaphlr{-}{35.8} & 79.9\cgaphlr{-}{10.7} & 57.4\cgaphlr{-}{28.4} & -13.7 \\
        \rowcolor{lightgray}
        GraNd~\citep{paul2021deep}
        & 91.2\cgaphlr{-}{4.4} & 78.6\cgaphlr{-}{12.0} & 21.0\cgaphlr{-}{57.4} & 39.2\cgaphlr{-}{52.1} & 7.1\cgaphlr{-}{78.4} & 6.9\cgaphlr{-}{58.7} & 37.2\cgaphlr{-}{53.5} & 32.3\cgaphlr{-}{53.5} & -46.2 \\
        \rowcolor{lightgray}
        Glister~\cite{killamsetty2021glister}
        & 91.4\cgaphlr{-}{4.2} & 86.8\cgaphlr{-}{3.8} & 71.8\cgaphlr{-}{6.6} & 80.2\cgaphlr{-}{11.1} & 67.9\cgaphlr{-}{17.5} & 32.8\cgaphlr{-}{32.8} & 80.9\cgaphlr{-}{9.7} & 71.5\cgaphlr{-}{14.3} & -12.5 \\ 
        \rowcolor{lightgray}
        Moderate~\citep{xia2022moderate}
        & 90.8\cgaphlr{-}{4.8} & 88.2\cgaphlr{-}{2.4} & 66.9\cgaphlr{-}{11.4} & 89.9\cgaphlr{-}{1.4} & 48.8\cgaphlr{-}{36.6} & 17.4\cgaphlr{-}{48.2} & 80.1\cgaphlr{-}{10.5} & 57.7\cgaphlr{-}{28.1} & -17.9 \\ 
        \rowcolor{lightgray}
        SSP~\citep{sorscher2022beyond}
        & 91.5\cgaphlr{-}{4.1} & 87.1\cgaphlr{-}{3.6} & 65.8\cgaphlr{-}{12.6} & 81.2\cgaphlr{-}{10.1} & 50.7\cgaphlr{-}{34.8} & 16.3\cgaphlr{-}{49.4} & 79.4\cgaphlr{-}{11.2} & 58.6\cgaphlr{-}{27.2} & -19.1 \\
        \rowcolor{lightgray}
        Prune4ReL$^\dagger$~\citep{park2023robust}
        & 91.6\cgaphlr{-}{4.1} & 85.9\cgaphlr{-}{4.8} & 63.9\cgaphlr{-}{14.5} & 78.4\cgaphlr{-}{12.9} & 49.6\cgaphlr{-}{35.9} & 16.3\cgaphlr{-}{49.3} & 78.8\cgaphlr{-}{11.8} & 56.7\cgaphlr{-}{29.1} & -20.3 \\
        \rowcolor{lightgray}
        Dyn-Unc~\citep{he2024large}
        & 95.4\cgaphlr{-}{0.2} & 89.3\cgaphlr{-}{1.4} & 78.8\cgaphl{+}{0.5} & 90.9\cgaphlr{-}{0.4} & 84.5\cgaphlr{-}{1.0} & 47.1\cgaphlr{-}{18.5} & 80.7\cgaphlr{-}{10.0} & 71.2\cgaphlr{-}{14.6} & -5.7 \\
        \rowcolor{lightgray}
        DUAL~\citep{cho2025lightweight}
        & 95.4\cgaphlr{-}{0.2} & 90.1\cgaphlr{-}{0.6} & 81.3\cgaphl{+}{3.4} & 92.0\cgaphl{+}{0.7} & 83.6\cgaphlr{-}{1.8} & 41.8\cgaphlr{-}{23.8} & 89.0\cgaphlr{-}{1.7} & 75.3\cgaphlr{-}{10.5} & -4.3 \\
        Dynamic Random~\cite{okanovic2024repeated}  
        & 94.5\cgaphlr{-}{1.1} & 89.2\cgaphlr{-}{1.5} & 76.1\cgaphlr{-}{2.3} & 88.9\cgaphlr{-}{2.4} & 83.7\cgaphlr{-}{1.8} & 63.3\cgaphlr{-}{2.3} & 90.6\cgap{+}{0.0} & 83.7\cgaphlr{-}{2.1} & -1.7 \\
        DivBS~\citep{hong2024diversified}
        & 94.9\cgaphlr{-}{0.7} & 89.8\cgaphlr{-}{0.8} & 78.0\cgaphlr{-}{0.4} & 85.9\cgaphlr{-}{5.4} & 79.6\cgaphlr{-}{5.9} & 48.6\cgaphlr{-}{17.0} & 88.8\cgaphlr{-}{1.8} & 84.6\cgaphlr{-}{1.2} & -4.2 \\
        IES~\citep{yuan2025instance}
        & 95.0\cgaphlr{-}{0.6} & 89.8\cgaphlr{-}{0.8} & 78.6\cgaphl{+}{0.2} & 87.0\cgaphlr{-}{4.3} & 79.1\cgaphlr{-}{6.4} & 35.7\cgaphlr{-}{29.9} & 88.5\cgaphlr{-}{2.2} & 78.3\cgaphlr{-}{7.5} & -6.4 \\
        InfoBatch~\citep{qin2024infobatch}
        & 95.0\cgaphlr{-}{0.6} & 90.5\cgaphlr{-}{0.2} & 77.7\cgaphlr{-}{0.6} & 92.0\cgaphl{+}{0.7} & 87.5\cgaphl{+}{2.1} & 64.3\cgaphlr{-}{1.3} & 92.0\cgaphl{+}{1.3} & 87.5\cgaphl{+}{1.7} & +0.4 \\
        \textbf{InfoBatch + Ours}  
        & \cellcolor{cyan!20}95.0\cgaphlr{-}{0.6} & \cellcolor{cyan!20}91.3\cgaphl{+}{0.6} & \cellcolor{cyan!20}82.4\cgaphl{+}{4.1} & \cellcolor{cyan!20}92.8\cgaphl{+}{1.5} & \cellcolor{cyan!20}87.6\cgaphl{+}{2.1} & \cellcolor{cyan!20}64.5\cgaphlr{-}{1.1} & \cellcolor{cyan!20}92.1\cgaphl{+}{1.5} & \cellcolor{cyan!20}89.8\cgaphl{+}{4.0} & \cellcolor{cyan!20}\textbf{+1.5} \\
        SeTa~\citep{zhou2025scale}
        & 94.8\cgaphlr{-}{0.8} & 89.4\cgaphlr{-}{1.3} & 78.5\cgaphl{+}{0.2} & 91.3\cgap{+}{0.0} & 86.0\cgaphl{+}{0.5} & 53.0\cgaphlr{-}{12.7} & 91.7\cgaphl{+}{1.1} & 87.9\cgaphl{+}{2.1} & -1.4 \\
        \textbf{SeTa + Ours} 
        & \cellcolor{cyan!20}94.9\cgaphlr{-}{0.7} & \cellcolor{cyan!20}89.7\cgaphlr{-}{1.0} & \cellcolor{cyan!20}79.3\cgaphl{+}{1.0} & \cellcolor{cyan!20}91.9\cgaphl{+}{0.6} & \cellcolor{cyan!20}87.6\cgaphl{+}{2.2} & \cellcolor{cyan!20}60.2\cgaphlr{-}{5.5} & \cellcolor{cyan!20}92.2\cgaphl{+}{1.6} & \cellcolor{cyan!20}88.5\cgaphl{+}{2.7} & \cellcolor{cyan!20}\textbf{+0.1} \\
        \midrule
        \multicolumn{10}{c}{\textbf{Prune Ratio $\sim$70\%}} \\
        \midrule
        \rowcolor{lightgray}
        Static Random 
        & 90.2\cgaphlr{-}{5.4} & 86.6\cgaphlr{-}{4.1} & 69.0\cgaphlr{-}{9.3} & 85.3\cgaphlr{-}{6.0} & 73.8\cgaphlr{-}{11.6} & 41.4\cgaphlr{-}{24.3} & 82.3\cgaphlr{-}{8.3} & 75.3\cgaphlr{-}{10.5} & -9.9 \\
        \rowcolor{lightgray}
        SmallL~\citep{jiang2018mentornet}
        & 75.1\cgaphlr{-}{20.5} & 76.3\cgaphlr{-}{14.3} & 74.4\cgaphlr{-}{4.0} & 78.2\cgaphlr{-}{13.1} & 76.1\cgaphlr{-}{9.4} & 35.4\cgaphlr{-}{30.2} & 73.1\cgaphlr{-}{17.6} & 62.7\cgaphlr{-}{23.1} & -16.5 \\
        \rowcolor{lightgray}
        Margin~\citep{coleman2019selection}
        & 91.2\cgaphlr{-}{4.4} & 71.7\cgaphlr{-}{19.0} & 15.1\cgaphlr{-}{63.3} & 22.6\cgaphlr{-}{68.7} & 12.3\cgaphlr{-}{73.1} & 9.8\cgaphlr{-}{55.8} & 34.0\cgaphlr{-}{56.7} & 27.3\cgaphlr{-}{58.5} & -49.9 \\
        \rowcolor{lightgray}
        Forget~\citep{toneva2018empirical}
        & 89.8\cgaphlr{-}{5.8} & 86.2\cgaphlr{-}{4.5} & 70.9\cgaphlr{-}{7.5} & 85.5\cgaphlr{-}{5.8} & 71.5\cgaphlr{-}{13.9} & 33.0\cgaphlr{-}{32.7} & 78.4\cgaphlr{-}{12.2} & 55.2\cgaphlr{-}{30.6} & -14.1 \\
        \rowcolor{lightgray}
        GraNd~\citep{paul2021deep}
        & 78.1\cgaphlr{-}{17.5} & 52.1\cgaphlr{-}{38.6} & 10.2\cgaphlr{-}{68.1} & 14.3\cgaphlr{-}{77.0} & 8.2\cgaphlr{-}{77.2} & 7.8\cgaphlr{-}{57.8} & 18.9\cgaphlr{-}{71.8} & 28.5\cgaphlr{-}{57.3} & -58.1 \\
        \rowcolor{lightgray}
        Glister~\cite{killamsetty2021glister}
        & 88.5\cgaphlr{-}{7.1} & 83.7\cgaphlr{-}{7.0} & 65.1\cgaphlr{-}{13.3} & 75.0\cgaphlr{-}{16.4} & 58.3\cgaphlr{-}{27.1} & 37.1\cgaphlr{-}{28.6} & 75.3\cgaphlr{-}{15.4} & 64.7\cgaphlr{-}{21.1} & -17.0 \\
        \rowcolor{lightgray}
        Moderate~\citep{xia2022moderate}
        & 87.8\cgaphlr{-}{7.8} & 84.8\cgaphlr{-}{5.8} & 66.3\cgaphlr{-}{12.0} & 87.3\cgaphlr{-}{4.0} & 47.8\cgaphlr{-}{37.7} & 16.7\cgaphlr{-}{48.9} & 77.7\cgaphlr{-}{13.0} & 55.4\cgaphlr{-}{30.4} & -19.9 \\ 
        \rowcolor{lightgray}
        SSP~\citep{sorscher2022beyond}
        & 87.8\cgaphlr{-}{7.8} & 82.0\cgaphlr{-}{8.7} & 60.5\cgaphlr{-}{17.8} & 73.9\cgaphlr{-}{17.4} & 41.2\cgaphlr{-}{44.2} & 15.8\cgaphlr{-}{49.9} & 74.7\cgaphlr{-}{15.9} & 55.1\cgaphlr{-}{30.7} & -24.0 \\
        \rowcolor{lightgray}
        Prune4ReL$^\dagger$~\citep{park2023robust}
        & 88.4\cgaphlr{-}{7.2} & 82.2\cgaphlr{-}{8.5} & 59.7\cgaphlr{-}{18.6} & 73.3\cgaphlr{-}{18.0} & 43.5\cgaphlr{-}{42.0} & 16.2\cgaphlr{-}{49.4} & 74.2\cgaphlr{-}{16.5} & 56.1\cgaphlr{-}{29.7} & -23.7 \\
        \rowcolor{lightgray}
        Dyn-Unc~\citep{he2024large}
        & 92.2\cgaphlr{-}{3.4} & 86.3\cgaphlr{-}{4.3} & 79.6\cgaphl{+}{1.2} & 90.1\cgaphlr{-}{1.2} & 86.8\cgaphl{+}{1.3} & 47.2\cgaphlr{-}{18.5} & 81.9\cgaphlr{-}{8.8} & 71.4\cgaphlr{-}{14.4} & -6.0 \\
        \rowcolor{lightgray}
        DUAL~\citep{cho2025lightweight}
        & 93.5\cgaphlr{-}{2.1} & 90.3\cgaphlr{-}{0.3} & 80.4\cgaphl{+}{2.1} & 91.7\cgaphl{+}{0.4} & 86.6\cgaphl{+}{1.1} & 45.2\cgaphlr{-}{20.5} & 90.8\cgaphl{+}{0.1} & 75.2\cgaphlr{-}{10.6} & -3.7 \\
        Dynamic Random~\cite{okanovic2024repeated}  
        & 93.0\cgaphlr{-}{2.6} & 87.4\cgaphlr{-}{3.2} & 75.9\cgaphlr{-}{2.5} & 87.4\cgaphlr{-}{3.9} & 84.6\cgaphlr{-}{0.8} & 60.2\cgaphlr{-}{5.5} & 91.4\cgaphl{+}{0.8} & 86.8\cgaphl{+}{1.0} & -2.1 \\
        DivBS~\citep{hong2024diversified}
        & 94.8\cgaphlr{-}{0.8} & 89.8\cgaphlr{-}{0.9} & 79.8\cgaphl{+}{1.5} & 86.2\cgaphlr{-}{5.1} & 80.3\cgaphlr{-}{5.1} & 48.5\cgaphlr{-}{17.1} & 90.0\cgaphlr{-}{0.6} & 83.8\cgaphlr{-}{2.0} & -3.8 \\
        IES~\citep{yuan2025instance}
        & 94.5\cgaphlr{-}{1.1} & 89.0\cgaphlr{-}{1.6} & 76.7\cgaphlr{-}{1.6} & 86.1\cgaphlr{-}{5.3} & 74.6\cgaphlr{-}{10.9} & 35.4\cgaphlr{-}{30.3} & 87.6\cgaphlr{-}{3.0} & 73.3\cgaphlr{-}{12.5} & -8.3 \\
        InfoBatch~\citep{qin2024infobatch} 
        & 94.3\cgaphlr{-}{1.3} & 89.9\cgaphlr{-}{0.8} & 79.4\cgaphl{+}{1.0} & 92.4\cgaphl{+}{1.1} & 87.0\cgaphl{+}{1.6} & 61.0\cgaphlr{-}{4.7} & 92.5\cgaphl{+}{1.8} & 88.3\cgaphl{+}{2.5} & +0.2 \\
        \textbf{InfoBatch + Ours} 
        & \cellcolor{cyan!20}94.3\cgaphlr{-}{1.3} & \cellcolor{cyan!20}90.8\cgaphl{+}{0.1} & \cellcolor{cyan!20}81.0\cgaphl{+}{2.7} & \cellcolor{cyan!20}92.8\cgaphl{+}{1.5} & \cellcolor{cyan!20}87.9\cgaphl{+}{2.5} & \cellcolor{cyan!20}61.1\cgaphlr{-}{4.5} & \cellcolor{cyan!20}92.9\cgaphl{+}{2.3} & \cellcolor{cyan!20}89.2\cgaphl{+}{3.4} & \cellcolor{cyan!20}\textbf{+0.8} \\
        SeTa~\citep{zhou2025scale} 
        & 94.3\cgaphlr{-}{1.3} & 89.5\cgaphlr{-}{1.2} & 77.7\cgaphlr{-}{0.6} & 92.2\cgaphl{+}{0.9} & 87.1\cgaphl{+}{1.7} & 30.4\cgaphlr{-}{35.2} & 92.2\cgaphl{+}{1.5} & 88.1\cgaphl{+}{2.3} & -4.0 \\
        \textbf{SeTa + Ours} 
        & \cellcolor{cyan!20}94.5\cgaphlr{-}{1.1} & \cellcolor{cyan!20}90.0\cgaphlr{-}{0.7} & \cellcolor{cyan!20}79.4\cgaphl{+}{1.1} & \cellcolor{cyan!20}92.6\cgaphl{+}{1.3} & \cellcolor{cyan!20}87.7\cgaphl{+}{2.3} & \cellcolor{cyan!20}58.0\cgaphlr{-}{7.6} & \cellcolor{cyan!20}92.5\cgaphl{+}{1.9} & \cellcolor{cyan!20}88.5\cgaphl{+}{2.7} & \cellcolor{cyan!20}\textbf{+0.0} \\
        \bottomrule
    \end{tabular}
    \vspace{-6pt}
    \label{tab:supp_results_cifar10_full}
\end{table*}

%% file: tab_supp/supp_results_cifar100_with_SOP.tex
\begin{table*}[t]
    \centering
    \footnotesize
    \setlength{\tabcolsep}{3pt}
    \caption{
    \textbf{Classification results of re-labeling integration on CIFAR-100N with ResNet-18.} 
    %
    We combine \algname{} with SOP+~\citep{liu2022robust} and compare it against re-labeling-augmented pruning baselines.
    Performance gaps relative to the full-training setting are indicated as superscripts.
    The mean $\Delta$ is computed across all noise types.
    }
    \vspace{-3pt}
    \begin{tabular}{lccccccc}
        \toprule
        Noisy Type $\rightarrow$ & \multirow{2}{*}{Real} & \multicolumn{3}{c}{Symmetric} & \multicolumn{2}{c}{Asymmetric} & Mean \\ 
        \cmidrule(lr){3-5}
        \cmidrule(lr){6-7}
        Pruning Method $\downarrow$ & & 0.2 & 0.5 & 0.8 & 0.2 & 0.4 & $\Delta$ \\
        \midrule
        Full-training 
        & 56.1 & 71.4 & 58.6 & 39.8 & 72.4 & 63.3 & -- \\
        \midrule
        \multicolumn{8}{c}{\textbf{Prune Ratio $\sim$30\%}} \\
        \midrule
        Prune4ReL~\citep{park2023robust} 
        & 64.4\cgaphl{+}{8.3} & 73.0\cgaphl{+}{1.7} & 68.2\cgaphl{+}{9.6} & 22.3\cgaphlr{-}{17.5} & 72.5\cgaphl{+}{0.1} & 65.9\cgaphl{+}{2.6} & +0.8 \\
        InfoBatch~\citep{qin2024infobatch}
        & 66.5\cgaphl{+}{10.4} & 74.0\cgaphl{+}{2.6} & 68.9\cgaphl{+}{10.3} & 21.5\cgaphlr{-}{18.3} & 74.6\cgaphl{+}{2.2} & 71.7\cgaphl{+}{8.4} & +2.6 \\
        \textbf{InfoBatch + Ours} 
        & \cellcolor{cyan!20}67.5\cgaphl{+}{11.4} & \cellcolor{cyan!20}76.3\cgaphl{+}{5.0} & \cellcolor{cyan!20}70.4\cgaphl{+}{11.8} & \cellcolor{cyan!20}32.5\cgaphlr{-}{7.3} & \cellcolor{cyan!20}75.8\cgaphl{+}{3.5} & \cellcolor{cyan!20}72.9\cgaphl{+}{9.6} & \cellcolor{cyan!20}\textbf{+5.7} \\
        \midrule
        \multicolumn{8}{c}{\textbf{Prune Ratio $\sim$50\%}} \\
        \midrule
        Prune4ReL~\citep{park2023robust} 
        & 62.2\cgaphl{+}{6.1} & 69.9\cgaphlr{-}{1.5} & 63.5\cgaphl{+}{4.9} & 23.8\cgaphlr{-}{16.0} & 68.0\cgaphlr{-}{4.4} & 61.1\cgaphlr{-}{2.1} & -2.2 \\
        InfoBatch~\citep{qin2024infobatch}
        & 66.8\cgaphl{+}{10.7} & 75.2\cgaphl{+}{3.8} & 69.0\cgaphl{+}{10.4} & 24.0\cgaphlr{-}{15.8} & 74.3\cgaphl{+}{1.9} & 71.6\cgaphl{+}{8.4} & +3.2 \\
        \textbf{InfoBatch + Ours}  
        & \cellcolor{cyan!20}67.6\cgaphl{+}{11.5} & \cellcolor{cyan!20}76.2\cgaphl{+}{4.9} & \cellcolor{cyan!20}71.6\cgaphl{+}{12.9} & \cellcolor{cyan!20}32.9\cgaphlr{-}{7.0} & \cellcolor{cyan!20}75.8\cgaphl{+}{3.4} & \cellcolor{cyan!20}73.1\cgaphl{+}{9.8} & \cellcolor{cyan!20}\textbf{+5.9} \\
        \midrule
        \multicolumn{8}{c}{\textbf{Prune Ratio $\sim$70\%}} \\
        \midrule
        Prune4ReL~\citep{park2023robust} 
        & 56.9\cgaphl{+}{0.7} & 62.9\cgaphlr{-}{8.5} & 57.1\cgaphlr{-}{1.5} & 15.1\cgaphlr{-}{24.7} & 62.5\cgaphlr{-}{9.8} & 54.6\cgaphlr{-}{8.7} & -8.7 \\
        InfoBatch~\citep{qin2024infobatch} 
        & 65.8\cgaphl{+}{9.7} & 74.5\cgaphl{+}{3.2} & 69.4\cgaphl{+}{10.8} & 20.5\cgaphlr{-}{19.3} & 74.2\cgaphl{+}{1.8} & 70.7\cgaphl{+}{7.4} & +2.3 \\
        \textbf{InfoBatch + Ours} 
        & \cellcolor{cyan!20}66.0\cgaphl{+}{9.9} & \cellcolor{cyan!20}76.4\cgaphl{+}{5.0} & \cellcolor{cyan!20}70.5\cgaphl{+}{11.9} & \cellcolor{cyan!20}30.7\cgaphlr{-}{9.1} & \cellcolor{cyan!20}75.2\cgaphl{+}{2.8} & \cellcolor{cyan!20}72.3\cgaphl{+}{9.0} & \cellcolor{cyan!20}\textbf{+4.9} \\
        \bottomrule
    \end{tabular}
    \vspace{-3pt}
    \label{tab:supp_results_cifar100_with_sop}
\end{table*}

%% file: tab_supp/supp_results_cifar10_with_SOP.tex
\begin{table*}[htbp]
    \centering
    \footnotesize
    \setlength{\tabcolsep}{3pt}
    \caption{
    \textbf{Classification results of re-labeling integration on CIFAR-10N with ResNet-18.} 
    %
    We combine \algname{} with SOP+~\citep{liu2022robust} and compare it against re-labeling-augmented pruning baselines.
    Performance gaps relative to the full-training setting are indicated as superscripts.
    The mean $\Delta$ is computed across all noise types.
    }
    \vspace{-3pt}
    \begin{tabular}{lcccccccc}
        \toprule
        Noisy Type $\rightarrow$ & \multicolumn{2}{c}{Real} & \multicolumn{3}{c}{Symmetric} & \multicolumn{2}{c}{Asymmetric} & Mean \\ 
        \cmidrule(lr){2-3}
        \cmidrule(lr){4-6}
        \cmidrule(lr){7-8}
        Pruning Method $\downarrow$ & Real-A & Real-W & 0.2 & 0.5 & 0.8 & 0.2 & 0.4 & $\Delta$ \\
        \midrule
        Full-training 
        & 90.7 & 78.3 & 91.3 & 85.4 & 65.6 & 90.6 & 85.8 & -- \\
        \midrule
        \multicolumn{9}{c}{\textbf{Prune Ratio $\sim$30\%}} \\
        \midrule
        Prune4ReL~\citep{park2023robust} 
        & 94.0\cgaphl{+}{3.4} & 89.7\cgaphl{+}{11.4} & 94.2\cgaphl{+}{2.9} & 91.1\cgaphl{+}{5.7} & 56.5\cgaphlr{-}{9.2} & 94.0\cgaphl{+}{3.4} & 92.0\cgaphl{+}{6.2} & +3.4 \\
        InfoBatch~\citep{qin2024infobatch}
        & 94.8\cgaphl{+}{4.1} & 91.1\cgaphl{+}{12.8} & 95.2\cgaphl{+}{3.9} & 94.0\cgaphl{+}{8.6} & 66.8\cgaphl{+}{1.2} & 94.7\cgaphl{+}{4.1} & 93.2\cgaphl{+}{7.4} & +6.0 \\
        \textbf{InfoBatch + Ours} 
        & \cellcolor{cyan!20}95.1\cgaphl{+}{4.4} & \cellcolor{cyan!20}91.7\cgaphl{+}{13.4} & \cellcolor{cyan!20}95.8\cgaphl{+}{4.5} & \cellcolor{cyan!20}94.6\cgaphl{+}{9.1} & \cellcolor{cyan!20}70.7\cgaphl{+}{5.1} & \cellcolor{cyan!20}95.5\cgaphl{+}{4.9} & \cellcolor{cyan!20}93.5\cgaphl{+}{7.8} & \cellcolor{cyan!20}\textbf{+7.0} \\
        \midrule
        \multicolumn{9}{c}{\textbf{Prune Ratio $\sim$50\%}} \\
        \midrule
        Prune4ReL~\citep{park2023robust} 
        & 92.6\cgaphl{+}{1.9} & 88.4\cgaphl{+}{10.1} & 92.8\cgaphl{+}{1.5} & 88.7\cgaphl{+}{3.3} & 49.5\cgaphlr{-}{16.2} & 92.9\cgaphl{+}{2.2} & 89.9\cgaphl{+}{4.1} & +1.0 \\
        InfoBatch~\citep{qin2024infobatch}
        & 94.9\cgaphl{+}{4.2} & 90.0\cgaphl{+}{11.7} & 95.0\cgaphl{+}{3.7} & 92.8\cgaphl{+}{7.4} & 66.8\cgaphl{+}{1.1} & 94.7\cgaphl{+}{4.1} & 92.0\cgaphl{+}{6.2} & +5.5 \\
        \textbf{InfoBatch + Ours}  
        & \cellcolor{cyan!20}95.0\cgaphl{+}{4.4} & \cellcolor{cyan!20}90.0\cgaphl{+}{11.7} & \cellcolor{cyan!20}96.0\cgaphl{+}{4.7} & \cellcolor{cyan!20}94.7\cgaphl{+}{9.2} & \cellcolor{cyan!20}71.1\cgaphl{+}{5.4} & \cellcolor{cyan!20}95.3\cgaphl{+}{4.6} & \cellcolor{cyan!20}93.2\cgaphl{+}{7.4} & \cellcolor{cyan!20}\textbf{+6.8} \\
        \midrule
        \multicolumn{9}{c}{\textbf{Prune Ratio $\sim$70\%}} \\
        \midrule
        Prune4ReL~\citep{park2023robust} 
        & 90.2\cgaphlr{-}{0.5} & 84.0\cgaphl{+}{5.7} & 90.2\cgaphlr{-}{1.1} & 83.3\cgaphlr{-}{2.2} & 42.8\cgaphlr{-}{22.8} & 89.8\cgaphlr{-}{0.8} & 84.0\cgaphlr{-}{1.8} & -3.3 \\
        InfoBatch~\citep{qin2024infobatch} 
        & 94.0\cgaphl{+}{3.3} & 87.7\cgaphl{+}{9.4} & 94.5\cgaphl{+}{3.2} & 91.3\cgaphl{+}{5.9} & 51.4\cgaphlr{-}{14.3} & 92.4\cgaphl{+}{1.7} & 90.4\cgaphl{+}{4.6} & +2.0 \\
        \textbf{InfoBatch + Ours} 
        & \cellcolor{cyan!20}94.0\cgaphl{+}{3.3} & \cellcolor{cyan!20}87.2\cgaphl{+}{8.8} & \cellcolor{cyan!20}95.1\cgaphl{+}{3.8} & \cellcolor{cyan!20}92.2\cgaphl{+}{6.8} & \cellcolor{cyan!20}58.5\cgaphlr{-}{7.2} & \cellcolor{cyan!20}94.8\cgaphl{+}{4.1} & \cellcolor{cyan!20}91.7\cgaphl{+}{5.9} & \cellcolor{cyan!20}\textbf{+3.7} \\
        \bottomrule
    \end{tabular}
    \vspace{-3pt}
    \label{tab:supp_results_cifar10_with_sop}
\end{table*}

%% file: tab_supp/supp_ablations_window_size.tex
\begin{table*}[htbp]
    \centering
    \footnotesize
    \setlength{\tabcolsep}{5pt}
    \caption{
    \textbf{Ablation on \algname{} with varying window sizes.} 
    We evaluate \algname{} using window sizes ranging from 2 to 50, aiming to validate the necessity of such trajectory window design.}
    \vspace{-3pt}
    \begin{tabular}{lccccccc}
        \toprule
        Noisy Type $\rightarrow$ & \multirow{2}{*}{Clean} & \multirow{2}{*}{Real} & \multicolumn{3}{c}{Symmetric} & \multicolumn{2}{c}{Asymmetric}\\ 
        \cmidrule(lr){4-6}
        \cmidrule(lr){7-8}
        Window Size $N$ $\downarrow$ & & & 0.2 & 0.5 & 0.8 & 0.2 & 0.4 \\
        \midrule
        \multicolumn{1}{c}{2}
        & 77.3 & 55.2 & 70.3 & 58.8 & 40.3 & 71.4 & 63.2 \\
        \multicolumn{1}{c}{3}
        & 78.0 & 55.9 & 70.7 & 59.2 & 40.8 & 71.2 & 63.1 \\
        \multicolumn{1}{c}{4 - 50 (avg ± std)}
        & 78.9\footnotesize{ ± 0.2} & 56.5\footnotesize{ ± 0.3} & 71.3\footnotesize{ ± 0.3} & 60.4\footnotesize{ ± 0.3} & 41.4\footnotesize{ ± 0.3} & 72.2\footnotesize{ ± 0.2} & 64.3\footnotesize{ ± 0.3} \\
        \bottomrule
    \end{tabular}
    \vspace{-3pt}
    \label{tab:supp_ablations_window_size}
\end{table*}

%% file: tab_supp/supp_results_pseudo_val_ablation.tex
\begin{table}[t]
    \centering
    \footnotesize
    \setlength{\tabcolsep}{3.6pt}
    \vspace{3pt}
    \caption{
    \textbf{More results with estimated pseudo clean set.}}
    \vspace{-3pt}
    \begin{tabular}{lcccccc}
        \toprule
        Reference-set Fraction $\rightarrow$ & \multicolumn{2}{c}{$100\%$} & \multicolumn{2}{c}{$10\%$} & \multicolumn{2}{c}{$1\%$} \\ 
        \cmidrule(lr){2-3}
        \cmidrule(lr){4-5}
        \cmidrule(lr){6-7}
        Reference-set Type $\downarrow$ & Clean & Real & Clean & Real & Clean & Real \\
        \midrule
        Clean
        & 79.0 & 56.8 & 78.9 & 56.3 & 79.1 & 56.0 \\
        Pseudo by ELFS~\cite{zheng2025elfs}
        & 78.8 & 56.6 & 79.1 & 56.3 & 78.9 & 56.1 \\
        \bottomrule
    \end{tabular}
    \vspace{-9pt}
    \label{tab:results_pseudo_val_more}
\end{table}

%% file: tab_supp/supp_results_statistic_analysis.tex
\begin{table*}[htbp]
    \centering
    \footnotesize
    \setlength{\tabcolsep}{5pt}
    \caption{
    \textbf{Statistical significance analysis on CIFAR-100N with ResNet-18.}}
    \vspace{-3pt}
    \begin{tabular}{lccccccc}
        \toprule
        Noisy Type $\rightarrow$ & \multirow{2}{*}{Clean} & \multirow{2}{*}{Real} & \multicolumn{3}{c}{Symmetric} & \multicolumn{2}{c}{Asymmetric}\\ 
        \cmidrule(lr){4-6}
        \cmidrule(lr){7-8}
        Pruning Method $\downarrow$ & & & 0.2 & 0.5 & 0.8 & 0.2 & 0.4\\
        \midrule
        InfoBatch
        & 79.0\footnotesize{ ± 0.1} & 56.1\footnotesize{ ± 0.2} & 71.4\footnotesize{ ± 0.2} & 59.7\footnotesize{ ± 0.2} & \textbf{41.8\footnotesize{ ± 0.1}} & 71.9\footnotesize{ ± 0.2} & 64.2\footnotesize{ ± 0.2} \\
        \textbf{InfoBatch + Ours} 
        & \cellcolor{cyan!20}\textbf{79.3\footnotesize{ ± 0.1}} & \cellcolor{cyan!20}\textbf{59.4\footnotesize{ ± 0.2}} & \cellcolor{cyan!20}\textbf{71.8\footnotesize{ ± 0.1}} & \cellcolor{cyan!20}\textbf{66.0\footnotesize{ ± 0.2}} & \cellcolor{cyan!20}\textbf{41.8\footnotesize{ ± 0.1}} & \cellcolor{cyan!20}\textbf{72.6\footnotesize{ ± 0.1}} & \cellcolor{cyan!20}\textbf{68.0\footnotesize{ ± 0.2}} \\
        \midrule
        SeTa
        & 79.0\footnotesize{ ± 0.1} & 55.6\footnotesize{ ± 0.1} & 70.2\footnotesize{ ± 0.1} & 59.0\footnotesize{ ± 0.2} & \textbf{41.6\footnotesize{ ± 0.1}} & 71.4\footnotesize{ ± 0.2} & 63.2\footnotesize{ ± 0.2} \\
        \textbf{SeTa + Ours}
        & \cellcolor{cyan!20}\textbf{79.3\footnotesize{ ± 0.1}} & \cellcolor{cyan!20}\textbf{56.3\footnotesize{ ± 0.1}} & \cellcolor{cyan!20}\textbf{70.8\footnotesize{ ± 0.2}} & \cellcolor{cyan!20}\textbf{60.5\footnotesize{ ± 0.2}} & \cellcolor{cyan!20}\textbf{41.6\footnotesize{ ± 0.1}} & \cellcolor{cyan!20}\textbf{71.9\footnotesize{ ± 0.1}} & \cellcolor{cyan!20}\textbf{64.3\footnotesize{ ± 0.2}} \\
        \bottomrule
    \end{tabular}
    \vspace{-6pt}
    \label{tab:supp_results_statistical_analysis}
\end{table*}

%% file: tab_supp/supp_results_news.tex
\begin{table}[!t]
    \centering
    \footnotesize
    \vspace{3pt}
    \setlength{\tabcolsep}{3.6pt}
    \caption{
    \textbf{Classification results on NEWS with 3-layer MLP.} 
    }
    \vspace{-3pt}
    \begin{tabular}{lcccc}
        \toprule
        Noisy Type $\rightarrow$ & \multirow{2}{*}{Clean} & \multicolumn{2}{c}{Symmetric} & \multicolumn{1}{c}{Pairflip}\\ 
        \cmidrule(lr){3-4}
        Pruning Method $\downarrow$ & & 0.2 & 0.5 & 0.45\\
        \midrule
        Full-training
        & 42.5 & 37.1 & 26.7 & 26.5\\
        InfoBatch
        & 42.8\cgaphl{+}{0.3} & 36.7\cgaphlr{-}{0.4} & 26.2\cgaphlr{-}{0.6} & 27.4\cgaphl{+}{0.9}\\
        \textbf{InfoBatch + Ours}
        & \cellcolor{cyan!20}\textbf{42.9}\cgaphl{+}{0.4} & \cellcolor{cyan!20}\textbf{37.9}\cgaphl{+}{0.8} & \cellcolor{cyan!20}\textbf{28.4}\cgaphl{+}{1.7} & \cellcolor{cyan!20}\textbf{27.7}\cgaphl{+}{1.2} \\
        \bottomrule
    \end{tabular}
    \vspace{-9pt}
    \label{tab:supp_results_news}
\end{table}